# Monitoring Teams by Overhearing: A Multi-Agent Plan-Recognition Approach

**Gal A. Kaminka** GALK@CS.BIU.AC.IL
*Computer Science Department*
*Bar Ilan University*
*Ramat Gan 52900, Israel*

**David V. Pynadath** PYNADATH@ISI.EDU
**Milind Tambe** TAMBE@USC.EDU
*Computer Science Department and Information Sciences Institute*
*University of Southern California*
*4676 Admiralty Way*
*Los Angeles, CA 90292, USA*

## Abstract

Recent years are seeing an increasing need for on-line monitoring of teams of cooperating agents, e.g., for visualization, or performance tracking. However, in monitoring deployed teams, we often cannot rely on the agents to always communicate their state to the monitoring system. This paper presents a non-intrusive approach to monitoring by *overhearing*, where the monitored team's state is inferred (via plan-recognition) from team-members' *routine* communications, exchanged as part of their coordinated task execution, and observed (overheard) by the monitoring system. Key challenges in this approach include the demanding run-time requirements of monitoring, the scarceness of observations (increasing monitoring uncertainty), and the need to scale-up monitoring to address potentially large teams. To address these, we present a set of complementary novel techniques, exploiting knowledge of the social structures and procedures in the monitored team: (i) an efficient probabilistic plan-recognition algorithm, well-suited for processing communications as observations; (ii) an approach to exploiting knowledge of the team's social behavior to predict future observations during execution (reducing monitoring uncertainty); and (iii) monitoring algorithms that trade expressivity for scalability, representing only certain useful monitoring hypotheses, but allowing for any number of agents and their different activities to be represented in a single coherent entity. We present an empirical evaluation of these techniques, in combination and apart, in monitoring a deployed team of agents, running on machines physically distributed across the country, and engaged in complex, dynamic task execution. We also compare the performance of these techniques to human expert and novice monitors, and show that the techniques presented are capable of monitoring at human-expert levels, despite the difficulty of the task.

## 1. Introduction

Recent years have seen tremendous growth of applications involving distributed multi-agent teams, formed of agents that collaborate on a specific joint task (e.g., Jennings, 1995; Pechoucek, Marik, & Stepankova, 2000, 2001; Kumar & Cohen, 2000; Kumar, Cohen, & Levesque, 2000; Horling, Benyo, & Lesser, 2001; Lenser, Bruce, & Veloso, 2001; Barber & Martin, 2001). This growth has led to increasing need for monitoring techniques that allow a synthetic agent or human operator to monitor and identify the state of the distributed team. Previous work has discussed the critical role of monitoring in visualization (e.g., Ndumu, Nwana, Lee, & Collis, 1999), in identifying failures in execution (e.g., Horling et al., 2001), in providing advice to improve performance (e.g., Aiello, Busetta, Dona, & Serafini, 2001), and in facilitating collaboration between the monitoring agent and the members of the team (e.g., Grosz & Kraus, 1996).

This paper focuses on monitoring cooperative agent teams by overhearing their internal communications. This allows a human operator or a synthetic agent to monitor the coordinated execution of a task, by listening to the messages team-members exchange with each other. It contrasts with previous techniques that are impractical in settings where direct observations of the team members are unavailable (e.g., when team-members are physically distributed away from the observer), or in large-scale applications composed of *already-deployed* agents that are dynamically integrated to jointly execute a task.

For example, one common technique, *report-based monitoring,* requires each monitored team-member to communicate its state to the monitoring agent at regular intervals, or at least whenever the team-member changes its state. Such reporting provides the monitoring agent with accurate information on the state of the team. Unfortunately, report-based monitoring suffers from several difficulties in monitoring large deployed teams of interest in the real-world (see Section 2 for a detailed discussion): First, it requires intrusive modifications to the behavior of agents, such that they report their state as needed by the different monitoring applications. However, since agents are already deployed, such repeated modifications to the behavior of the agents are difficult to implement and complex to manage. In particular, legacy and proprietary systems are notoriously expensive to modify (for instance, consider the notorious modifications to address the Year 2000 bug, also known as Y2K). Second, the bandwidth requirements of report-based monitoring (which relies on communication channels) can be unrealistic (Jennings, 1993, 1995; Grosz & Kraus, 1996; Pechoucek et al., 2000, 2001; Vercouter, Beaune, & Sayettat, 2000). In addition, network delays and unreliable or lossy communication channels are a key concern with report-based monitoring approaches.

We therefore advocate an alternative monitoring approach, based on multi-agent keyhole plan-recognition (Tambe, 1996; Huber & Hadley, 1997; Devaney & Ram, 1998; Intille & Bobick, 1999; Kaminka & Tambe, 2000). In this approach, the monitoring system infers the unobservable state of the agents based on their observable actions, using knowledge of the plans that give rise to the actions. This approach is non-intrusive, requiring no changes to agents' behaviors; and it allows for changes in the requested monitoring information. It assumes access to knowledge of plans that may explain observable action—however this knowledge is readily available to the monitoring system as we assume it is deployed in a collaborative environment. Indeed, in some cases, the monitoring system may be deployed by the human operator of the team. An additional benefit of a plan-recognition approach is that it can rely on inference to compensate for occasional communication losses, and can therefore be robust to communication failures.

In general, the only observable actions of agents in a distributed team are their *routine* communications, which the agents exchange as part of task execution (Ndumu et al., 1999). Fortunately, the growing popularity of agent integration tools (Tambe, Pynadath, Chauvat, Das, & Kaminka, 2000; Martin, Cheyer, & Moran, 1999) and agent communications (Finin,

Labrou, & Mayfield, 1997; Reed, 1998) increases standardization of aspects of agent communications, and provides increasing opportunities for observing and interpreting inter-agent communications. We assume that monitored agents are truthful in their messages, since they are communicating to their teammates; and that they are not attempting to deceive the monitoring agent or prevent it from overhearing (as it is deployed by the human operator of the team). Given a (possibly stochastic) model of the plans that the agents may be executing, a monitoring system using plan-recognition can infer the current state of the agents from such observed routine messages.

However, the application of plan-recognition techniques for overhearing poses significant challenges. First, a key characteristic of the overhearing task is the scarcity of observations. Explanations for overheard messages (i.e., the observed actions) can sometimes be fairly easy to disambiguate, but uncertainty arises because there are relatively few of them to observe: team members cannot and do not in practice continuously communicate among themselves about their state (Jennings, 1995; Grosz & Kraus, 1996). Thus team-members change their state while keeping quiet. Another key characteristic of overhearing is that the observable actions are inherently *multi-agent actions*: When agents communicate, it is only a single agent that *sends* the messages. The others implicitly act their role in the communications by *listening*. Yet despite the scarcity of observable communications, and the multi-agent nature of the observed actions, a monitoring system must infer the state of all agents in the team, at all times. Previous investigations of multi-agent plan-recognition (Tambe, 1996; Devaney & Ram, 1998; Intille & Bobick, 1999; Kaminka & Tambe, 2000) have typically made the assumption that all changes to the state of agents have an observable effect: Uncertainty resulted from ambiguity in the explanations for the observed actions. Furthermore, these investigations have addressed settings where observable actions were individual (each action is carried out by a single agent).

In addition to these challenges that are unique to overhearing, a monitoring system must address additional challenges stemming from the use of monitoring in service of visualization. The representation and algorithms must support soft real-time response; reasoning must be done quickly to be useful for visualization. Furthermore, real-world applications demand techniques that can scale up as the number of agents increases, for monitoring large teams. However, many current representations for plan-recognition are computationally intense (e.g., Kjærulff, 1992), or only address single-agent recognition tasks (e.g., Pynadath & Wellman, 2000). Multi-agent plan-recognition investigations have typically not explicitly addressed scalability concerns (Devaney & Ram, 1998; Intille & Bobick, 1999).

This paper presents Overseer, an implemented monitoring system capable of monitoring large distributed applications composed of previously-deployed agents. Overseer builds on previous work in multi-agent plan-recognition (Tambe, 1996; Intille & Bobick, 1999; Kaminka & Tambe, 2000) by utilizing knowledge of the relationships between agents to understand how their decisions interact. However, as previous techniques proved insufficient, Overseer includes a number of novel multi-agent plan-recognition techniques that address the scarcity of observations, as well as the severe response-time and scale-up requirements imposed by realistic applications. Key contributions include: (i) a *linear time* probabilistic plan-recognition representation and associated algorithms, which exploit the nature of observed communications for efficiency; (ii) a method for addressing unavailable observations by exploiting knowledge of the *social procedures* of teams to effectively predict (and hence effectively monitor) future observations during normal and failed execution, thus allowing inference from lack of such observations; and (iii) YOYO*, an algorithm that uses knowledge of the team organizational structure (*team-hierarchy*) to model the agent team (with all the different parallel activities taken by individual agents) using a single structure, instead of modeling each agent individually. YOYO* sacrifices some expressivity (the ability to accurately monitor the team in certain coordination failure states) for significant gains in efficiency and scalability.

We present a rigorous evaluation of Overseer's different monitoring techniques in one of its application domains and show that the techniques presented result in significant boosts to Overseer's monitoring accuracy and efficiency, beyond techniques explored in previous work. We evaluate Overseer's capability to address lossy observations, a key concern with report-based monitoring. Furthermore, we evaluate Overseer's performance in comparison with human expert and novice monitors, and show that Overseer's performance is comparable to that of human experts, despite the difficulty of the task, and Overseer's reliance on computationally-simple techniques. One of the key lessons that we draw in Overseer is that a combination of computationally-cheap multi-agent plan-recognition techniques, exploiting knowledge of the expected structures and interactions among team-members, can be competitive with approaches which focus on accurate modeling of individual agents (and may be computationally expensive).

This paper is organized as follows. Section 2 presents the motivation for the design of Overseer, using examples from an actual distributed application in which Overseer was applied. Section 3 presents a novel single-agent plan-recognition representation and associated algorithms, particularly suited to monitoring an agent based on its observed communications. Section 4 explores several methods Overseer uses to address uncertainty in using this representation for monitoring a team of agents. Section 5 presents YOYO*, which allows efficient reasoning using the methods previously discussed. Section 6 presents an evaluation of the different techniques incorporated in YOYO*. Section 7 contrasts the techniques presented with previous related investigations, and finally, Section 8 concludes and presents our plans for future work. In addition, several appendices present all pseudo-code for algorithms discussed in the text, and portions of the data used in our experiments, for those readers who may wish to replicate the experiments.

## 2. Motivation and Illustrative Examples

Several considerations, based on our experience with actual distributed applications, have directed us towards the plan-recognition approach we advocate in this paper. We present these considerations in the context of an illustrative complex distributed application, which we also use for evaluating Overseer in Section 6. In this application, a distributed team of 11 to 20 agents executes a simulation of an evacuation of civilians from a threatened location. The integrated system allows a human commander to interactively provide locations of the stranded civilians, safe areas for evacuation and other key points. Simulated helicopters then fly a coordinated mission to evacuate the civilians, relying on various information agents to dynamically obtain information about enemy threats, (re)plan routes to avoid threats and obstacles, etc. The distributed team is composed of diverse agents from four different research groups: A Quickset multi-modal command input agent (Cohen, Johnston, McGee,

Oviatt, Pittman, Smith, Chen, & Clow, 1997), a Retsina route planner (Payne, Sycara, Lewis, Lenox, & Hahn, 2000), the Ariadne information agent (Knoblock, Minton, Ambite, Ashish, Modi, Muslea, Philpot, & Tejada, 1998) and eight synthetic helicopter pilots (Tambe, Johnson, Jones, Koss, Laird, Rosenbloom, & Schwamb, 1995).

The agents were not designed to work together on this task—they were already built and deployed prior to the creation of the team. The team is integrated using Teamcore (Tambe et al., 2000), which accomplishes integration by "wrapping" each agent with a proxy that maintains collaboration with other agents (via their own proxies). The proxies and agents form a team, jointly executing a distributed application described by a *team-oriented program.* Such a program consists of:

- A team hierarchy, where a team decomposes into subteams, and sub-subteams.
- A plan hierarchy, which contains team plans that decompose into subteam plans
- Assignment of teams from the team hierarchy to plans in the plan hierarchy.

As an example, Figure 1-a shows a part of the team/subteam hierarchy used in the evacuation-domain (described below). Here, for instance, TRANSPORT is a subteam of FLIGHT-TEAM, itself a subteam of TASK-FORCE. Figure 1-b shows an abbreviated plan-hierarchy for the same domain. High-level team plans, such as `Evacuate`, typically decompose into other team plans, such as `Process-Orders`, and, ultimately, into leaf-level plans that are executed by individuals. Temporal transitions are used to constrain the order of execution of plans. There are teams assigned to execute the plans, e.g., the TASK FORCE team jointly executes Evacuate, while only the TRANSPORT subteam executes the Transport-Operations (Transport-Ops) step. The team-oriented program for this application consists of about 40 team-plans. Some plans may get executed repeatedly though, so each agent may execute up to hundreds of plan steps as part of the execution of a single team-oriented program.

To execute the team-oriented program, each proxy uses a domain-independent teamwork model, called STEAM (Tambe, 1997). The teamwork model automatically generates the communication messages required to ensure appropriate coordination among the proxies. For instance, STEAM requires that if an agent privately obtains a belief *bel* that terminates a team plan, then that agent should send a message to the rest of the team to terminate that team plan, along with the private belief *bel* that led to that termination. To avoid jamming the communication channels with a flood of messages about every single plan, STEAM chooses to communicate selectively. Thus, whereas communicating about the initiation and termination of each and every plan would have led to 2000 or more messages generated in one run, only about 100 messages get exchanged in any one run when using STEAM (Tambe et al., 2000).

Figure 2 displays some of the messages exchanged among team members in the evacuation application, through the use of STEAM. The first message is sent from a proxy called "teamquickset" to members of a team TEAM-EVAC (another name for TASK FORCE). The content of this message indicates that the team should terminate a plan called `determine-number-of-helos`. The second message is sent from a proxy called

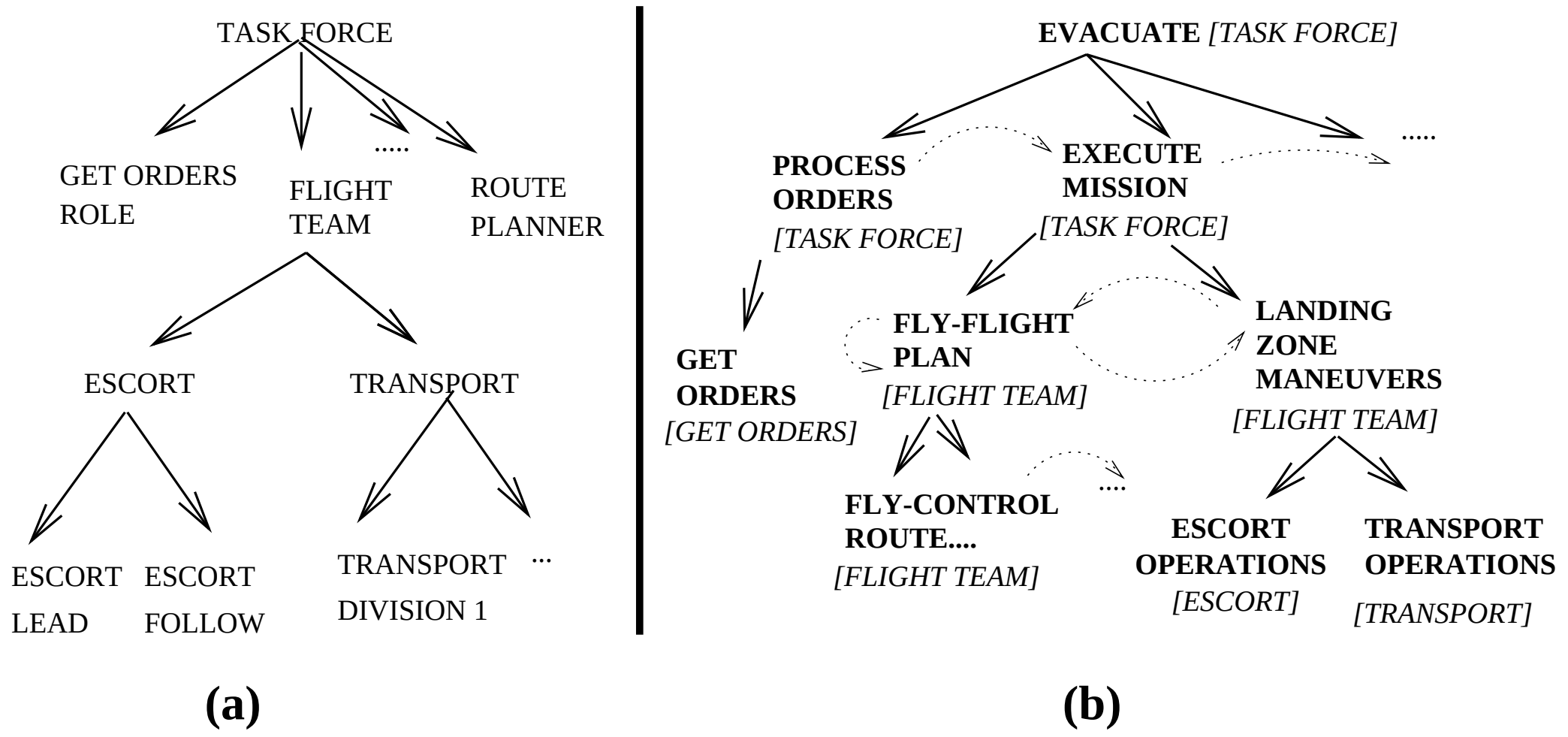

Figure 1: Portions of the team-hierarchy (a) and plan-hierarchy (b) used in our domain. Dotted lines show temporal transitions.

"team_auto2" to members of a subteam TEAM-ESCORT-FOLLOW (a subteam of ESCORTS). The content of this message indicates that the subteam should establish commitment to a plan named `prepare-to-execute-mission`. The online appendix presents sample logs of the overheard messages from complete runs, as well as the plan and team hierarchies for the evacuation application.

As discussed in Section 1, the capability for automatically monitoring the progress of the team is critical. This need for team monitoring is further amplified in distributed settings, since a human operator in one place cannot directly observe the agents executing in a remote location. For instance, in trial runs of the evacuation simulation application described above, monitoring sometimes required a series of frantic phone calls among human operators in different states, trying to verify the successful execution of the system as it was operating. And even when this agent team was co-located on multiple computers in one room, the diversity of agents made it extremely difficult for an observer to automatically monitor the state of the team just from observing the different agent output screens.

Overseer was built to provide such monitoring by tracking the routine communications among the agents (Figure 2). Using plan-recognition, it allows humans and agents to query about the present and future likely plans of the entire team, its subteams and individuals—to monitor progress, compute likelihoods of failure, etc. However, given that the agent team communicates selectively about the plans being executed, Overseer's plan-recognition faces significant uncertainty. Furthermore, Overseer must be able to answer queries on-line, and must therefore work efficiently. As discussed later, addressing these challenges has required several novel team-based plan-recognition techniques to be developed.

Several considerations have led us away from report-based monitoring for this and other Teamcore applications. First, report-based monitoring requires that agents' code be modified to communicate the reports needed for monitoring; as monitoring requirements change

```
Log Message Received; Fri Sep 17 18:27:54 1999:
   Logging Agent: teamquickset
   Message==> tell
   :content teamquickset terminate-jpg constant determine-number-of-helos
      number-of-helos-determined *yes* 4 4 98 kqml_string
   :receiver TEAM-EVAC 9 kqml_word
   :reply-with nil 3 kqml_word
   :team TEAM-EVAC 9 kqml_word
   :sender teamquickset 12 kqml_word
   :kqml-msg-id 21547+tsevet.isi.edu+7 22 kqml_word

 Log Message Received; Fri Sep 17 18:30:35 1999:
   Logging Agent: TEAM_auto2
   Message==> tell
   :content TEAM_auto2 establish-commitment prepare-to-execute-mission
      58 kqml_string
   :receiver TEAM-ESCORT-FOLLOW 18 kqml_word
   :reply-with nil 3 kqml_word
   :team TEAM-ESCORT-FOLLOW 18 kqml_word
   :sender TEAM_auto2 10 kqml_word
   :kqml-msg-id 20752+dui.isi.edu+16 20 kqml_word
```

Figure 2: Example KQML messages used as observations by Overseer.

from one application to the next, so does the information needed about each agent. Unfortunately, the agents and their proxies are already deployed in several government laboratories and universities. Modifying the agents at each deployed location is problematic and intrusive—modifications interfere with carefully designed timing specifications of given tasks, requiring further modifications by other agent developers. The distributed nature of Teamcore implies that there is no centralized server which controls the behavior of the agents, but instead changes are required in the different proxy types. Indeed, in general, modifying legacy and proprietary applications (including the integration architecture) is of course known to be a difficult process, and so a solution that requires constant modifications to the agents and architecture will not scale up.

A second important consideration was the computational and bandwidth requirements of report-based monitoring. As has been repeatedly noted in the literature, one cannot expect agents to be able to communicate continuously and fully monitor all other agents (e.g., Jennings, 1993, 1995; Grosz & Kraus, 1996; Pechoucek et al., 2001; Vercouter et al., 2000). In a team of 11 (used as an example in this paper), regularly scheduled state reports from the agents at the required temporal resolution would require approximately 50,000 messages to be sent during a 15-minute run, with the number nearly doubling when we reach 20 agents. If we instead have the 11 agents only report on state changes, announcing plan initiation and termination, approximately 2,000 messages have to be sent. However, this is still an order-of-magnitude more than the normal 100 messages or so that are exchanged by the 11 agents as part of routine execution. Even if the network could support the bandwidth necessary for report-based monitoring, there is also a significant computational burden on the monitoring system to process all the incoming reports.

On the other hand, a plan-recognition approach seemed like a natural fit for the task. First, it doesn't require any changes in the behavior of the monitored agents, and is thus very suitable for monitoring agents that are already deployed. Second, it doesn't add any computational burdens to the monitored agents or the network, since it uses only what observations are already available. Third, the main knowledge source plan-recognition systems typically rely on—a plan library—is in fact easily available in accessible form to the monitoring system from the team-oriented program which is used to integrate the agents, since the operator deploying the monitoring system is assumed to be the one to describe the integration team-oriented program in the first place. Thus plan-recognition's sometimes criticized assumption of a correct plan-library is in fact satisfied fully in this monitoring application.

Note that this assumption holds even if agents are not all using the same integration architecture: The only knowledge we rely on is a (possibly stochastic) model of how components of execution fit together, and the communications that are used to integrate them. Therefore, while this paper focuses on team-oriented programs (described above), the techniques introduced appear generalizable to other types of representation languages for distributed systems, such as TÆMS (Decker, 1995), team-oriented programming (Tidhar, 1993a) and others. Furthermore, the plan-library need not contain implemetation details—only the names of the key steps. Thus even agents utilizing radically-different representations than a plan-hierarchy can be monitored, as long as they have execution states corresponding to the team-oriented program (which they have to have in any case in order to coordinate with other team-members).

Monitoring by overhearing poses unique challenges as previously discussed. However, it also offers unique opportunities for plan recognition. We had earlier stated our assumption that agents are truthful in their communications, and do not seek to deceive their teammates or the monitoring system, nor prevent overhearing in any way (e.g., encryption). This assumption is justified as the monitoring system is deployed by the operator of the monitored agents, or by an agent team-member. Failures of the team to coordinate (e.g., due to clock asynchrony or unintentional erroneous messages) will therefore cause corresponding failures in monitoring. However, we do not make additional assumptions about the messages beyond those that are made by the monitored agents themselves.

This assumption allows a plan-recognition system to treat observations with certainty: When a message is overheard terminating plan $X$, the monitoring system can infer with certainty that indeed the plan $X$ is no longer executed. However, this does not eliminate plan recognition ambiguity. First, multiple instantiations of plan $X$ may exist, and the message does not specify which one was terminated. Second, upon termination of the plan, the monitored team-member must often choose between multiple alternative plan steps to follow $X$, and yet this choice is not evident in the observations. Indeed, the difficulty of monitoring by overhearing is demonstrated by human monitoring performance: Novice human monitors have managed to only achieve approximately 60% accuracy on average.

## 3. Monitoring a Team of Agents as Separate Individuals

In this section, we present a representation and associated baseline algorithms to support overhearing based on the plan-hierarchy and team-hierarchy. We begin by making an assumption of agent independence, where observations and beliefs about one agent's state of execution have no bearing on our beliefs about another agent's state. This assumption can be contrasted with another: If we assume instead that team-members are successful in their coordination, then knowing that one agent has begun executing a joint plan would naturally increase the likelihood that its teammates have begun as well, as agents would not be considered independent. In fact, successful teamwork *requires* interdependency among the agents (Grosz, 1996).

However, an initial assumption of agent independence provides a baseline of comparison, as it more closely follows current approaches to multi-agent plan recognition, which often assume that observations about each individual agent are continuously available. Later sections (Sections 4 and 5) will highlight the unique challenges tackled in monitoring by overhearing, and will take agent interdependencies into account.

We thus begin by maintaining a separate plan recognizer for each agent. Each recognizer observes only those messages that its respective agent sends. On the basis of these observations, the recognizer maintains a probabilistic estimate of the state of execution of the various plans the agent may be currently executing. Knowledge of the plans assigned to agents and their team memberships is available in our application from the plan-hierarchy and team-hierarchy of the team-oriented program used in constructing the monitored application.

Section 3.1 presents the language we use for the probabilistic representation of a team-oriented program. We exploit various independence properties within team-oriented programs to achieve a compact representation of the possible plan states of the agents. Sec-tion 3.2 presents an algorithm for updating the recognizer's beliefs about the agents' plan states upon the observation of a message. This algorithm performs the update with an efficiency gained by exploiting the particular semantics of communicated messages, namely that each such message is an observation that indicates the initiation/termination of a particular plan *with certainty.* Section 3.3 presents an algorithm for updating the recognizer's beliefs about the agents' plan states when *no* message has been observed. In the absence of any such evidence, this algorithm efficiently updates the recognizer's beliefs by using a temporal model of the agents' plan execution that makes a strong Markovian assumption. Finally, Section 3.4 presents the overall recognition procedure, as well as an illustration and complexity analysis of that procedure.

### 3.1 Plan-State Representation

We address uncertainty in monitoring through a probabilistic model that supports quantitative evaluation of the recognized plan hypotheses. Since we are monitoring these agents through the duration of their execution, we use a time series of plan-state variables. At each point in time, the agent's plan state is the state of the team-oriented program that it is currently executing, i.e., a path from root to leaf in the team-oriented program tree. We represent the plans in the program by a set of boolean random variables, $\{X_t\}$, where each variable $X_t$ is true if and only if the agent is actively executing plan $X$ at time $t$. We then represent our beliefs about the agent's actual state at time $t$ as a probability distribution over all variables $\{X_t\}$. The distribution takes into account dependencies among the different plans in the team-oriented program (e.g., parent-child relationships), as well as the temporal dependencies between the plan state at times $t$ and $t+1$. To simplify the dependency structure, it is useful to introduce additional boolean random variables, $done(X, t)$, that are true if and only if plan $X$ was executed at time $t-1$ and its execution has terminated at time $t$.

There are a number of possible representations for capturing the distribution and performing inference over these variables. However, the generality of the plan hierarchy, the dynamic nature of the domain, and the requirements of the task eliminate most existing approaches from consideration. For instance, we could potentially generate a DBN—Dynamic Belief Network (Kjærulff, 1992)—to represent the probabilistic distribution over the plan variables. To do so, we include nodes representing all of the plan variables, $X_t$, as well as representing $done(X, t)$. The links among these nodes represent the structure of the plan hierarchy (e.g., parent-child relationships, temporal constraints), and we can fill in the conditional probability tables accordingly. We also represent the temporal progress of the team by including nodes for the variables at the next time slice, $X_{t+1}$. We add links from the $X_t$ nodes to the $X_{t+1}$ nodes and represent the dynamics in the conditional probability tables on those links. For each transition from a node $X_t$ to a node $Y_{t+1}$ ($X \neq Y$), we would also add binary nodes indicating the observation of a message along that transition. Thus, for a plan hierarchy with $M$ plan nodes, the corresponding DBN representation will have $O(4M + M^2) = O(M^2)$ binary random variables.

The standard DBN inference algorithms maintain a belief state, $b_t$, representing the posterior probability distribution over the variables in time slice, $t$, conditioned on all of the observations made so far (from time 0–$t$). These inference algorithms can update the belief state to incorporate new evidence about any variables, $X_t$, and they can also compute the next time-tick's belief state, $b_{t+1}$. We can extract the desired probability over plan-state variables by examining the posterior probabilities stored in $b_t$. Given the dependency structure of our plan model, the space and time complexity of performing inference using this DBN (either incorporating a single observation, or computing $b_{t+1}$) is $O(2^{M^2})$ for a single agent.

This DBN method is not sufficiently efficient to support on-line monitoring in real-world domains, since on each and every time step, the recognizer must perform an inferential step of exponential computational complexity. There exist *single-agent* plan-recognition techniques that avoid the exponential complexity of DBNs by using a representation and inference algorithms aimed at the particular properties of the plan-recognition task (e.g., Pynadath & Wellman, 2000). Such specialized representations avoid the full generality of DBNs, while still capturing a broad class of interesting planning agent models. Given a specialized representation, the single-agent plan-recognition algorithms can exploit the particular structure of the plan models to achieve efficient online inference.

Drawing our inspiration from the success of this work in single-agent domains, we adopt a similar methodology in our multi-agent domain. In other words, we have developed a novel plan-recognition representation more suited to capturing team-oriented programs. The structural assumptions we make in this representation support efficient inference with our specialized algorithms, as well as more naturally supporting an extension to represent inter-agent dependencies (as discussed in Section 4).

We represent the team-oriented plan as a directed graph, whose vertices are plans, and whose edges signify temporal and hierarchical decomposition transitions between plans: Children edges denote hierarchical decomposition of a plan into sub-plans. Sibling edges denote temporal orderings between plans. Following the structure of the plan hierarchy, the variables $\{X_t\}$ form a directed connected graph, such that each node $X_t$ has at most one *hierarchical-decomposition incoming transition* from a parent node (representing its parent plan), and any number of *temporal incoming transitions* from plans that precede it in order of execution. The graph may contain multiple nodes for a single plan, if the plan is the potential child of multiple parent plans. The node may have any number of temporal outgoing transitions to immediate successor sibling nodes (representing plans that may follow it in order of execution), and any number of hierarchical-decomposition outgoing transitions to the node's *first* children (i.e., those that will be executed first by a decomposition of the plan $X_t$. The graph forms a tree along hierarchical decomposition transitions, so that no plan can have itself as a descendent. On the other hand, there may be cycles along temporal transitions (to siblings). In other words, a plan may have an outgoing temporal transition to itself (meaning that it can be selected for execution again upon termination), or to a node that has a temporal path leading back to the plan (meaning that it is the first node in a temporal sequence of plans that may be executed repeatedly). It may also have two alternative temporal paths leading indirectly from one node to another.

To perform inference with this representation, we borrow the standard DBN inference algorithms' notion of a belief state, $b_t$. As in the DBN case, the belief state represents the posterior probability distribution over the variables in time slice, $t$, conditioned on all of the observations made so far. In addition, for each plan, we distinguish between a state of actual execution and a *blocked* state, indicating that execution has terminated, but execution of a successor has not yet begun (perhaps because the agent is in the process of sending a message). Thus, $b_t(X, block)$ is our belief that $X$ has terminated, but the agent has not begun execution of a successor; $b_t(X, \neg block)$ is then our belief at time $t$ that the monitored agent is currently executing $X$, which has not yet terminated. More precisely, we define $b_t(X, block) \equiv \Pr(X_t, done(X, t+1)|\mathcal{E})$ and $b_t(X, \neg block) \equiv \Pr(X_t, \neg done(X, t+1)|\mathcal{E})$, where $\mathcal{E}$ again denotes all of the evidence we have received so far. If the recognizer observes a message from an agent at time $t$, it updates its previous belief state, $b_t$, by incorporating the evidence into its new belief state, $b_{t+1}$, according to the method described in Section 3.2. If it does not observe a message from an agent at time $t$, it propagates belief into its new belief state, $b_{t+1}$, using the method described in Section 3.3 to simulate plan execution over time.

### 3.2 Belief Update with Observed Message

While observing team communications, the recognizer can expect to occasionally receive evidence in the form of messages (sent by an individual agent member) that identify either plan initiation or termination. In incorporating this evidence, we exploit the assumption that the agents are truthful in their messages. In other words, if we observe an initiation message for a plan, $X$, at time $t$, then $X_t$ is true with certainty. Likewise, if we observe a termination message for a plan, $X$, at time $t$, then $done(X, t+1)$ is true with certainty. More precisely, the algorithms presented in this section are specialized to exploit the property of observed communications, where for any observation $\Omega$, either $\Pr(X_t|\Omega, \mathcal{E}) = 1$ or $\Pr(done(X, t)|\Omega, \mathcal{E}) = 1$, for any possible previously observed evidence, $\mathcal{E}$.

Though messages are assumed truthful, there still remains ambiguity. First, while a message uniquely specifies the relevant *plan*, it does not uniquely specify the relevant *node*. In other words, the recognizer is still unsure about which particular $X_t$ node the message refers to, since the graph may contain multiple $X_t$ nodes consistent with the message. Furthermore, when a message announces termination of a plan (even with no ambiguity about the corresponding node), there still remains ambiguity about the next plan selected by the agent.

The observations available in the overhearing tasks of immediate interest to us fall into this level of ambiguity. In our evacuation scenario example, there are two nodes corresponding to the plan `land-troops`, because there is one instance of `land-troops` for picking up the people to be transported and another for dropping them off. If the recognizer observes a message indicating that an agent has initiated execution of `land-troops`, then there is ambiguity about which of the two instances is currently relevant. Furthermore, there may exist ambiguity about which plan the agent will select after terminating `land-troops`.

Algorithm 1 presents the pseudo-code for the complete procedure for incorporating evidence from observations.

**Incorporating Evidence of an Observed Initiation Message (lines 3–8)** Suppose that, at time $t$, we have observed a message, *msg*, that corresponds to initiation. If only one plan, $X$, is consistent with *msg*, then we know, with certainty, that the agent is executing $X$, regardless of whatever evidence we have previously observed. Therefore, we can simply set our belief that $X_t$ is true to be 1.0. If multiple plans are consistent with *msg*, we distribute the unit probability over each consistent plan, weighted by our prior belief in seeing the given

**Algorithm 1** Incorporate-Evidence(**msg** $m$, **beliefs** $b$, **plans** $M$)

Initialize distributions $b', b_{t+1} \leftarrow 0.0$ for all plans in $M$
**for all** plans $X \in M$ consistent with $m$ **do**
  **if** $m$ is an initiation message **then**
    **for all** plans $W$ that precede $X$ **do**
      $b'(X, \neg block) \leftarrow b'(X, \neg block) + b_t(W, block)\mu_{wx}\pi_{wx}$
  **else** $\{m$ is a termination message$\}$
    **for all** plans $Y \in M$ that succeed $X$ **do**
      $b'(Y, \neg block) \leftarrow b'(Y, \neg block) + b_t(X, block)\mu_{xy}\pi_{xy}$
Normalize distribution $b'$
**for all** plans $X \in M$ with $b' > 0$ **do**
  $b_{t+1}(X, \neg block) \leftarrow b'(X, \neg block)$
  Propagate-Down$(X, b'(X, \neg block), b, M)$
  $tmp \leftarrow X$
  **while** $parent(tmp) \neq null$ **do**
    $b_{t+1}(parent(tmp), \neg block) \leftarrow b_{t+1}(parent(tmp), \neg block) + b_{t+1}(tmp, \neg block)$
    $tmp \leftarrow parent(tmp)$

message. This prior belief depends on all predecessor plans of $X$ that may have terminated prior to seeing this message.

To support the computation of the beliefs over transitions from predecessor plans to successors, as well as the beliefs of seeing a message for a given transition, Overseer stores two parameters: $\pi$ and $\mu$. The former is the probability of entering a successor plan, $X$, given that predecessor plan, $W$, has just completed: $\pi_{wx} \equiv \Pr(X_{t+1}|W_t, done(W, t+1))$. The latter is the probability of seeing a message, given that the agent took the specified transition: $\mu_{wx} \equiv \Pr(msg_t|W_t, done(W, t+1), X_{t+1})$. We can use previous runs to acquire suitable values for these parameters, $\pi$ and $\mu$, by producing a frequency count over transitions and messages seen during those runs (see Section 4.2 for more discussion of the use of $\mu$ in Overseer).

Therefore, given the observation of an initiation message, *msg*, at time $t$, we wish to distribute the unit probability over all plans, $X$, (in the *unblocked* state) that are consistent with *msg*. We can derive our new belief in plan $X$ at time $t+1$ as follows:

$$\Pr(X_{t+1}|msg, \mathcal{E}) = \frac{\Pr(msg, X_{t+1}|\mathcal{E})}{\Pr(msg|\mathcal{E})}$$

The denominator is simply a normalization factor, and it is the same for all candidate plans, $X$. Therefore, we ignore it in this derivation, and focus on only the numerator, which we can expand over all possible predecessor plans, $W$, and possible termination states of $W$:

$$\begin{aligned} &\propto \sum_W \Pr(msg, X_{t+1}, W_t, done(W, t+1)|\mathcal{E}) \\ &\quad + \sum_W \Pr(msg, X_{t+1}, W_t, \neg done(W, t+1)|\mathcal{E}) \end{aligned}$$

The second term is 0, since we cannot proceed from $W$ to $X$ if $W$ has *not* terminated. In the second term, we can expand the joint probability into its component conditional probabilities:

$$\propto \sum_W [\Pr(msg|W_t, done(W, t+1), X_{t+1}, \mathcal{E}) \\ \cdot \Pr(X_{t+1}|W_t, done(W, t+1), \mathcal{E}) \\ \cdot \Pr(W_t, done(W, t+1)|\mathcal{E})]$$

We assume that the probability of sending a message and the distribution over plan transitions obey a Markov property, so that they are independent of the plan history before time $t$, given the current plan at time $t$. Thus, the first two conditional probabilities are independent of our previous history of observations. The third is exactly our previous belief that $W$ is blocked:

$$\propto \sum_W [\Pr(msg|W_t, done(W, t+1), X_{t+1}) \Pr(X_{t+1}|W_t, done(W, t+1)) \\ \cdot b_t(W, block)]$$

The first two conditional probabilities are exactly our parameters, $\mu$ and $\pi$:

$$\propto \sum_W \mu_{wx} \pi_{wx} b_t(W, block) \tag{1}$$

Lines 4–5 of Algorithm 1 perform exactly the derived summation of Equation 1 (the normalization step is carried out on line 9 (see below). A similar procedure is followed when a message is observed indicating the termination of $X$ (lines 6–8). In such a case, we know that the agent was executing $X$ in the previous time step but that it has moved on to some successor. Thus, for each of $X$'s potential successor plans $Y$, we set our belief in $Y$ to be proportional to a transition probability, similar to that for the initiation message:

$$\Pr(Y_{t+1}|msg, \mathcal{E}) = \frac{\Pr(msg, Y_{t+1}|\mathcal{E})}{\Pr(msg|\mathcal{E})}$$

The denominator is again a normalization factor that we ignore. We can expand the numerator over possible states of $X$'s execution:

$$\propto \Pr(msg, Y_{t+1}, X_t, done(X, t+1)|\mathcal{E}) \\ + \Pr(msg, Y_{t+1}, \neg X_t, done(X, t+1)|\mathcal{E}) \\ + \Pr(msg, Y_{t+1}, X_t, \neg done(X, t+1)|\mathcal{E}) \\ + \Pr(msg, Y_{t+1}, \neg X_t, \neg done(X, t+1)|\mathcal{E})$$

Only the first term is nonzero, since the others correspond to states of execution that are inconsistent with the observed message:

$$\propto \Pr(msg, Y_{t+1}, X_t, done(X, t+1) | \mathcal{E})$$

We can rewrite this joint probability as a product of conditional probabilities:

$$\begin{aligned} &\propto \Pr(msg | X_t, done(X, t+1), Y_{t+1}, \mathcal{E}) \\ &\quad \cdot \Pr(Y_{t+1} | X_t, done(X, t+1), \mathcal{E}) \\ &\quad \cdot \Pr(X_t, done(X, t+1) | \mathcal{E}) \end{aligned}$$

We again use our Markovian assumptions to simplify the conditional probabilities, and we rewrite the third probability using our belief state:

$$\begin{aligned} &\propto \Pr(msg | X_t, done(X, t+1), Y_{t+1}) \Pr(Y_{t+1} | X_t, done(X, t+1)) \\ &\quad \cdot b_t(X, block) \end{aligned}$$

Finally, we rewrite the first two conditional probabilities using our parameters, $\mu$ and $\pi$:

$$\propto \mu_{xy} \pi_{xy} b_t(X, block) \tag{2}$$

Lines 7–8 of Algorithm 1 perform exactly the derived summation of Equation 2.

**Normalization of the sum (line 9).** Line 9 normalizes the sum to recapture a well-formed probability distribution. Note that the normalization step must take into account the fact that evidence may be incorporated for plan steps where one is an ancestor of another—in which case the evidence for the ancestor plan is probabilistically redundant. The more specific evidence (for the descendent plan) will be more useful for visualization, as it is more accurate.

**Propagation of Evidence (lines 10–16)** Finally, the recalculated beliefs are set (line 11) and then the changes are recursively propagated down the decomposition hierarchy to the plan's children (line 12), via the call to Algorithm 2. In addition, the recalculated beliefs are propagated up to the plan's ancestors in the decomposition hierarchy (lines 13–16), since evidence of a child plan being active is evidence of its parent being active as well. We assume here that we have no knowledge about the relative likelihood of the child plans, so we treat each as equally likely. If we had additional knowledge about these likelihoods, we could easily exploit it in our Propagate-Down algorithm.

**Algorithm 2** Propagate-Down(**plan** $Y$, **probability** $\rho$, **beliefs** $b$, **plans** $M$)

$C \leftarrow \{c \mid c \in M,\ c \text{ first child of } Y\}$
$\rho' \leftarrow \rho / \mid C \mid$
**for all** plans $c \in C$ **do**
  $b_{t+1}(Y, \neg block) \leftarrow b_{t+1}(Y, \neg block) + \rho'$
  Propagate-Down$(c, \rho', b, M)$

### 3.3 Belief Update with No Observation

In overhearing tasks, there is a great deal of uncertainty about when agents complete the execution of their plan steps, since agents do not necessarily send messages upon every termination or initiation of a plan. Therefore, if no messages are observed at time $t$, then the system's beliefs for time $t+1$ must be calculated based on the possibility that the agents may have initiated or terminated plans without sending any messages. To support the necessary belief update, we need a model of plan execution that provides us with a probability of plan termination over time (i.e., $\Pr(done(X,t))$). In principle, this probability distribution can be arbitrarily complex, and its structure may vary enormously from domain to domain, and even from plan to plan within the same domain. In some domains, obtaining an accurate model of this distribution requires complex knowledge acquisition from domain experts or else a complex learning process on the part of the agent. In addition, an accurate model may be too complex to support efficient online inference.

Overseer instead uses a temporal model that supports both efficient inference and simple parameter estimation procedures. Overseer models the duration of a (leaf) plan, $X$, as an exponential random variable. In other words, the probability of the plan completing execution within $\tau$ time units increases as $1-e^{-\tau\lambda_X}$. The single parameter, $\lambda_X$, corresponds to $1/(\text{mean duration of } X)$, which we can easily acquire from domain experts or previous runs. As for inference, the exponential random variable has a Markovian property, in that the probability of the plan's completion between times $t$ and $t+1$ is

$$\Pr(done(X,t+1)|X_t) \equiv 1 - e^{-\lambda_x},$$

*independent* of how long the agent has been executing $X$ before time $t$. This strong assumption may not fully hold in some real-world domains, but it is often a good approximation. Also, the error associated with this approximation may be acceptable, given the enormous gain in inferential efficiency (as we show in the remainder of this section).

These efficiency gains manifest themselves when Overseer rolls the model forward in time to compute its belief state for the next time slice. Given the exponential random variable as a model of plan duration, the probability of completion of a leaf plan is a constant, $1 - e^{-\lambda_x}$, for each plan $X$. For plans with children, the probability of completion is exactly the probability of completion of its last child (according to the temporal ordering of the children).

Having computed the probability of plan termination, Overseer then evaluates which plan the agent may execute next. It examines the possible successors and, for each, computes the probability of taking the corresponding transition, conditioned on the fact that no message was observed $(1 - \mu_{xy})$, and on the prior probability of taking this message $(\pi_{xy})$. Again, as mentioned in Section 3.2, Overseer makes a Markovian assumption that the plan history before time $t$ does not affect the likelihood of the various transitions. Given this assumption, it can combine the two parameters, $\pi$ and $\mu$, to get the desired conditional probability of the transition, given that we observed no message:

$$
\begin{aligned}
&\Pr(Y_{t+1}|X_t, done(X, t+1), \neg msg_t) \\
=&\frac{\Pr(\neg msg_t|X_t, done(X, t+1), Y_{t+1}) \Pr(Y_{t+1}|X_t, done(X, t+1))}{\Pr(\neg msg_t|X_t, done(X, t+1))} \\
=&\frac{(1-\mu_{xy})\pi_{xy}}{\sum_Z \Pr(\neg msg_t|X_t, done(X, t+1), Z_{t+1}) \Pr(Z_{t+1}|X_t, done(X, t+1))} \\
=&\frac{(1-\mu_{xy})\pi_{xy}}{\sum_Z (1-\mu_{xz})\pi_{xz}} \\
=&\frac{(1-\mu_{xy})\pi_{xy}}{\eta_X}
\end{aligned} \tag{3}
$$

The normalizing denominator, $\eta_X$, is the sum of the numerator over all possible successors, $Y$, which we can pre-compute off-line. We can use the value of $\eta_X$ to determine the likelihood that the agent will send a message upon terminating plan $X$ at time $t$. In the special case when $\eta_X = 0$, Equation 3 is not well-defined, as *all* possible transitions from $X$ *require* a message. In this case, the agent cannot have begun execution of any successor, even though it has completed execution of $X$. $\eta_X$ is therefore the probability mass signifying our belief that the agent is no longer executing $X$ at time $t+1$, and is not waiting for a message (i.e., it is in a blocked state). In other words, it is our increased belief that the agent is executing one of $X$'s immediate successors at time $t+1$, given that we have seen no message.

Algorithm 3 presents the pseudo-code for the process of propagating the probabilities forward in time when a message is not observed. First, it initializes all the values to 0 (lines 1–5). The process continues by going over all plans $X \in M$, *in post-order*—we explore children plans (i.e., plans reachable by hierarchical decomposition transitions) before their parents, and sibling plans in order of execution. For each plan, the algorithm executes four stages: (1) It determines the plan's outgoing probabilities (lines 7–10); (2) it determines $\eta_x$, the outgoing probability mass that is propagated along the outgoing temporal transitions without being blocked by waiting for a message (lines 11–12); (3) it propagates $\eta_x$ along the non-blocked temporal outgoing transitions (lines 13–20); and finally (4) it computes our belief that the agent will execute the plan at the next time-tick $b_{t+1}(X, \neg block)$ or will be blocking (lines 21–22). The remainder of this section explains these four stages in detail.

**Calculating the outgoing probability $out_x$ (lines 7–10).** In Algorithm 3, the variable $out_x$ represents the total temporal outgoing probability from plan, $X$, given our belief that the agent was executing $X$ at time $t$. If a plan $X$ is a leaf, then we derive its temporal outgoing probability, $out_x$, from the temporal model discussed previously, given our belief that the agent is currently executing $X$ (lines 7–8). If $X$ is a parent, lines 9–10 are, in fact, redundant: They serve only to remind the reader that for a parent, $Y$, $out_y$ follows from $Y$'s children when they execute line 20. This depends critically on the post-order traversal of the plan-hierarchy: the outgoing probability of a parent $Y$ is derived from the outgoing probabilities of its last hierarchical-decomposition children, and thus all children's outgoing probabilities must be calculated before their parents'.

**Algorithm 3** Propagate-Forward(**beliefs** $b$, **plans** $M$)

1: **for all** plans $X \in M$ **do**
2: $\quad b_{t+1}(X, \neg block) \leftarrow 0.0$
3: $\quad b_{t+1}(X, block) \leftarrow 0.0$
4: $\quad out_x \leftarrow 0.0$
5: $\quad \eta_x \leftarrow 0.0$
6: **for all** plans $X \in M$ in post-order **do** {children in temporal order before parents}
7: $\quad$ **if** X is a leaf **then**
8: $\quad\quad out_x \leftarrow b_t(X, \neg block)(1 - e^{-\lambda_x})$ {calculate probability of $X$ terminating at time $t$}
9: $\quad$ **else** {X is a parent}
10: $\quad\quad out_x$ is known { because post-order guarantees all children set it in line 20}
11: $\quad$ **for all** *temporal* outgoing transitions $T_{x \to y}$ from $X$ **do**
12: $\quad\quad \eta_x \leftarrow \eta_x + (1 - \mu_{xy})\pi_{xy}$
13: $\quad$ **if** $\eta_x > 0$ **then** {some transition can be taken}
14: $\quad\quad$ **for all** temporal outgoing transitions $T_{x \to y}$ from $X$ **do**
15: $\quad\quad\quad \rho \leftarrow out_x(1 - \mu_{xy})\pi_{xy}$
16: $\quad\quad\quad$ **if** $T_{x \to y}$ leads to a successor plan $Y$ **then**
17: $\quad\quad\quad\quad b_{t+1}(Y, \neg block) \leftarrow b_{t+1}(Y, \neg block) + \rho$
18: $\quad\quad\quad\quad$ Propagate-Down$(Y, \rho, b, M)$
19: $\quad\quad\quad$ **else** {$T_{x \to y}$ is a terminating transition}
20: $\quad\quad\quad\quad out_{parent(x)} \leftarrow out_{parent(x)} + (1 - \mu_{xy})\pi_{xy}$ {parent's outgoing probability is its children's}
21: $\quad b_{t+1}(X, block) \leftarrow b_{t+1}(X, block) + out_x - \eta_x$
22: $\quad b_{t+1}(X, \neg block) \leftarrow b_{t+1}(X, \neg block) - out_x$

**Determining the non-blocked outgoing probability $\eta_x$ (lines 11–12).** The probability, $\eta_x$ is the sum over all possible values of the numerator in Equation 3 (i.e., over all temporal outgoing transitions originating in $X$), as illustrated in the derivation. As we see in line 21, $\eta_x$ is critical for calculating the belief that the agent has terminated execution of $X$, but has not yet begun execution of a successor (i.e., the belief $b_{t+1}(X, block)$ that the agent is blocking).

**Propagating $\eta_x$ along temporal outgoing transitions (lines 13–20).** This is the key component in the propagation. For every temporal outgoing transition $T_{x \to y}$, Overseer calculates $\rho$, a temporary variable that holds the probability mass corresponding to Overseer's belief in the joint event of (i) the agent having completed execution of $X$, (ii) the agent taking the transition $T_{X \to Y}$, and (iii) the agent doing so without sending out an observable message. The calculation of $\rho$ is derived as follows:

$$\begin{aligned}
\rho &= \text{Probability that } X \text{ is done } \wedge \text{ no message was observed } \wedge \text{ agent chose } T_{x \to y} \\
&= \Pr(done(X,t)|X_t)\Pr(\neg msg|X_t, done(X,t))\Pr(Y_{t+1}|X_t, done(X,t), \neg msg_t) \\
&= out_x * \eta_x * \frac{(1-\mu_{xy})\pi_{xy}}{\eta_x} \\
&= out_x * (1-\mu_{xy})\pi_{xy} \qquad (4)
\end{aligned}$$

If the transition $T_{x \to y}$ leads to a successor plan $Y$ (lines 16–18), then $\rho$ is added to $Y$'s future state (at time $t+1$) as temporal incoming probability. Since decomposition is assumed to be immediate, this incoming probability is propagated (added) to $Y$'s first children (Algorithm 2). If there are multiple first children, then they denote alternative plan decompositions for a single agent, and we compute the probability over them by dividing the probability incoming to the parent among them. If any children have first child plans of their own, we distribute this new incoming probability in turn, using the same method. Only in the next time-step does the algorithm propagate from first children to the next child, in order of execution. The reason for this is that we assume that all plans take at least a single time step to complete.

If the transition $T_{x \to y}$ is the special-case termination transition (line 19–20), then $X$ has no successors. In this case, the outgoing temporal probability $\rho$ is added to $X$'s parent's outgoing probability $out_{parent(x)}$ so that it may be used when propagating $parent(x)$'s temporal outgoing probability along its own temporal outgoing transitions. Note again that the post-order traversal of the plan-hierarchy guarantees that all children are explored before their parents, thus $out_{parent(x)}$ is fully computed by the time the algorithm reaches $parent(x)$.

**Computing $X$'s new blocked and non-blocked probabilities (lines 21–22).** Now that the outgoing probability mass has been propagated to $X$'s children and siblings, the only steps remaining involve re-calculation of OVERSEER's belief in $X$'s blocked and non-blocked states. The total temporal outgoing probability (whether blocked or not) is $out_x$; it must be subtracted from future belief that the agent is executing $X$. The probability mass that left $b_t(X, \neg block)$ but is blocking on a message that was not observed by OVERSEER is $out_x - \eta_x$. It is added to $X$'s future blocked state.

### 3.4 Discussion

The overhearing approach outlined in this section maintains a separate plan-recognition mechanism for each agent, ignoring any inter-agent dependencies. Using an array of individual models (Figure 3) that are updated with the passage of time, or as messages are observed, the state of a team is taken to be the combination of the most likely state of each individual agent. Algorithm 4 embodies this approach: It is called every time tick, collects all messages that are observed, and updates the state of the agents.

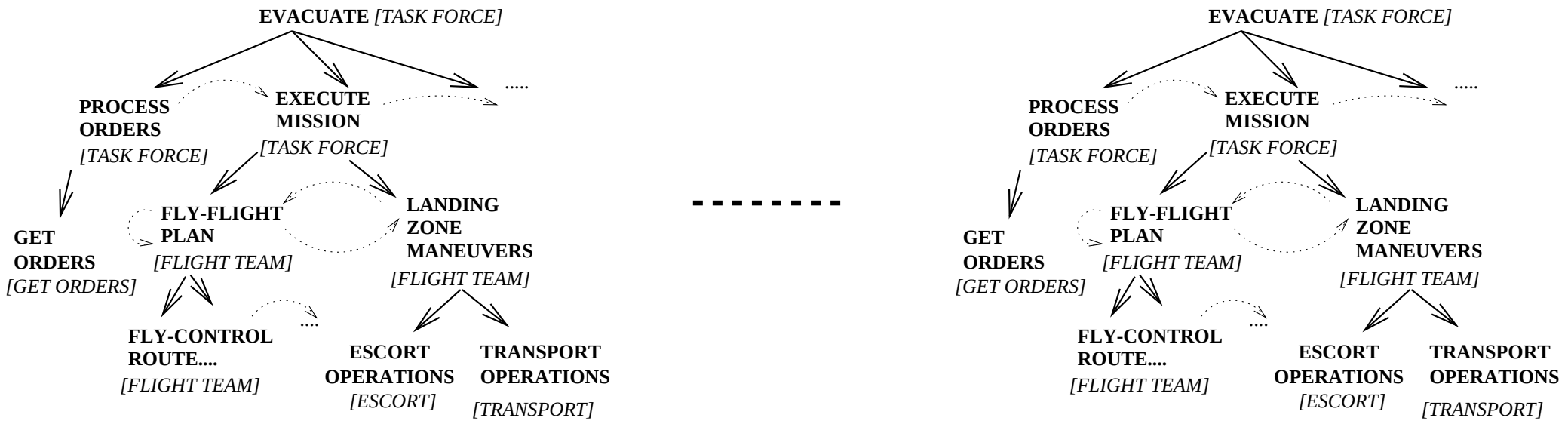

Figure 3: Array of single-agent recognizers—one for each agent.

As an illustration of the operation of this algorithm, consider the example domain of the evacuation scenario. OVERSEER begins with a belief that the agent is executing its top-level plan (and its first child, `Process-Orders`) at time 0 (i.e., $b_0(Evacuate, \neg block) = 1.0$, $b_0(ProcessOrders, \neg block) = 1.0$). If OVERSEER observes a message about the initiation of `Fly-Flight-Plan` by one of the helicopters, then it applies INCORPORATE-EVIDENCE (Algo-

**Algorithm 4** ARRAY-OVERSEER(**beliefs** $b$, **plan-hierarchy array** $M[]$, **agents** $A$)

```
for all Agents a ∈ A do
  if A message m_a from a was observed then
    INCORPORATE-EVIDENCE(m_a, b, M[a])
  else {No message was sent by a }
    PROPOGATE-FORWARD(b, M[a])
```

rithm 1). From the plan-hierarchy (Figure 1b) it is known that `Process-Orders` cannot be a possible current or future plan of the agent, and that the helicopter in question is executing `Fly-Flight-Plan`, i.e., $b_t(ProcessOrders, \neg block) = 0$, $b_t(FlyFlightPlan, \neg block) = 1.0$. This probability mass is propagated to `Fly-Flight-Plan`'s first children, of which there is one, and thus the belief in this child is set to 1.0 as well.

After some time passes and no message is observed, there is uncertainty as to whether `Fly-Flight-Plan` and `Landing-Zone-Maneuvers` are active, as both are possible future states, and the duration of `Fly-Flight-Plan` is uncertain. OVERSEER would still assign a probability of 1.0 to the top-level plan `Evacuate`. However, some probability mass from `Fly-Flight-Plan` would be propagated every time-tick to `Landing-Zone-Maneuvers` by PROPAGATE-FORWARD (Algorithm 3). For each such propagation, the incoming temporal probability mass being added to the belief in the execution of `Landing-Zone-Maneuvers` would be propagated to its first children immediately. Assuming that the helicopter agent is free to select either `Transport-Operations` or `Escort-Operations`, the incoming probability would be split evenly and added to the prior belief in each of the two first children. In the same temporal propagation step, any outgoing belief from these first children would be propagated via their own outgoing temporal transitions.

The inference procedure described by Algorithms 1–4 exploits the particular structure of our representation in ways that more general existing algorithms cannot. The pseudo-code demonstrates that for a single monitored agent, both types of belief updates have a time complexity *linear* in the number of plans and transitions in $M$, i.e., $O(M)$. Thus for $N$ agents, the space and time complexity of Algorithm 4 is $O(MN)$.

We gain this efficiency (compared to an approach such as DBN) from two sources. First, we make a Markovian assumption that the probability of observing a message depends on only the relevant plan being active, independently of execution history. With this assumption, we can incorporate evidence, based on only our beliefs at time $t$. Second, we make another Markovian assumption in the temporal model, allowing our propagation algorithm to reason forward to time $t+1$ based on only our beliefs at time $t$, without regard for previous history.

## 4. Monitoring a Team by Overhearing

The previous section has outlined an efficient plan-recognition mechanism that is particularly suitable for monitoring a single agent based on its communications. Monitoring a team was achieved by monitoring each member of the team independently of the others. Unfortunately, although the time complexity of this approach is acceptable, its monitoring (recognition) results are poor. The evaluation in Section 6.1 provides more details, but, in short, the average accuracy using this approach over all experiments was *less than 4%*.

The main cause for this low accuracy is the scarcity of observations, one of the identifying characteristics of monitoring by overhearing. As previously discussed, agents often switch their state unobservably (i.e., without sending a message). Therefore, the monitoring system critically needs to estimate correctly the times at which agents switch state. Since some agents rarely communicate (i.e., there are very few observations about them), variance in their temporal behavior (with respect to the system's predictions) tends to cause large errors in monitoring.

To address this issue, we bring back for discussion the agent independence assumption which we have made in the previous section. After all, team-members do not communicate independently of each other: Communication in a team is an action that is intended to change the state of a listener (Cohen & Levesque, 1990). Agents that only rarely *send* a message may still change their state upon *receiving* a message. In other words, although observed messages are used in the previous section to update the belief in the state of the sender, they could also be used to update the state of any listeners. To do this, the monitoring system must know about the relationships between the team-members.

Knowledge of the social structures enables additional sophisticated forms of monitoring. For instance, in order to maintain their social structures, team-members communicate with each other predictably, during particular points in the execution of a task. Such predictions of future observable behavior—communications—can be used to further reduce the uncertainty. However, it is often the case that while it can be difficult to correctly predict that a specific agent will communicate at a specific point in task execution, it is easy to predict that some team-member will. Knowledge of the procedures employed by a team to maintain its social structures can be very useful allows a monitoring system to make such predictions.

To reason about the effects of communications on receivers, and about future observable behavior of team-members, a monitoring system must utilize knowledge of the social structures and social procedures used by team-members to maintain these structures. Such exploitation of social knowledge for monitoring is called Socially-Attentive Monitoring (Kaminka & Tambe, 2000). This section discusses these concepts in detail.

### 4.1 Exploiting Social Structures

While computationally cheap, the approach described earlier proved insufficient in the evacuation domain. In monitoring by overhearing tasks, the monitoring system must address scarce observations, as agents rarely communicate all at the same time. Indeed, in the evacuation application, only a single message was observed (on average) for every 20 combined individual state changes.

Under such challenging conditions, a system for monitoring by overhearing must come to rely extensively on its ability to estimate when agents change their internal state without sending a message. The representation presented earlier used a simple, but efficient, temporal model to do this, based on the estimated average duration of plans. However, we have found high variance in the actual duration of plan execution, compared to the duration predicted by the average-duration model:

- Plan execution times vary depending on the *external environment*. For instance, when all the agents in the team are running on a local network, their response times to queries may be shorter than when communicating across continents. Indeed, latency times in the Internet vary greatly, and are difficult to predict.

- Plan execution times vary depending on *when a plan-step is executed internally.* For instance, the `traveling` plans, used repeatedly within the given evacuation team-oriented program, take anywhere from 15 seconds to almost two minutes to execute, depending on the particular route being followed.

- Plan execution times vary depending on *the outcome of a plan-step.* For instance, when the route-planner is functioning correctly, it responds within a few seconds. However, when it crashes it does not return an answer at all, and the other agents wait for a relatively long time before relying on a time-out to decide that it had failed.

This problem can be addressed in principle by a more expressive model of execution duration, for instance taking into account the internal execution context. However, in practice, such a model would likely be much more expensive computationally, as it would need to rely on knowledge of previous and future steps, breaking the Markovian assumption (e.g., to determine duration based on *when a plan-step is executed*, an improved temporal model would have to reason about the likelihood that a given instance of the plan-step is the second instance, as opposed to a third). As applications grow in scale in the real world, an increasingly more complex temporal model would have to be continuously refined to cover the increasingly complex temporal behavior of agents. Fortunately, a temporal model is only one way in which a monitoring system can estimate the times in which agents change their internal state unobservedly.

An alternative method for estimating unobserved state changes is to utilize known dependencies between agents to exploit evidence about the state of one agent to infer the state of another. In particular, it is often true in team settings that one agent would send a message *intending* to affect the state of all its receivers in a particular way. Thus in principle, under the assumption that the receivers do change their state predictably, an observation of such a message can be used as evidence in the inference of the sender's state, as well as all receivers', i.e., the state of all team-members. We can trade the agent independence assumption made earlier with an assumption of successful coordination. This is a reasonable assumption in team settings, given that agents are actively attempting to maintain their teamwork with such communications (Tambe, 1997; Kumar et al., 2000; Dunin-Keplicz & Verbrugge, 2001).

The effects of a message on a receiver are dependent on the relationship between the sender and the receiver (where we take such a relationship to be described by a mathematical relation between the possible states of the sender and the receiver). In principle, such relationships underly *social structures*—structures of interactions between agents that make the decisions of one team-member dependent, to some predictable degree, on those of its teammates. Using knowledge of these dependencies, a monitoring agent may use observations of a communication action by an agent to infer the possible state of another.

One simple example of such a structure is common in many teams (e.g., Jennings, 1993; Kinny, Ljungberg, Rao, Sonenberg, Tidhar, & Werner, 1992), and indeed is present also in our application: *roles* that govern which team-members undertake what tasks in service of the team goal. Such roles ideally bias the decision mechanism of the team-members towards making decisions that are appropriate for their roles. Thus knowledge of the roles of team-members can be useful to counter the uncertainty faced by a monitoring agent. For instance, suppose the monitoring agent knows that in the evacuation application, a particular team-member is to choose `Transport-Ops`, rather than `Escort-Ops`, as a child of `Landing-Zone-Maneuvers` (because the team-member belongs to the TRANSPORT team, rather than the ESCORT team). This knowledge can reduce the uncertainty the monitoring agent has—under the assumption that the team-member did not incorrectly choose an *inappropriate* plan for its role. Overseer in fact uses knowledge of roles in such a manner to alleviate uncertainty. This monitoring use of role information has been used in previous work (Tambe, 1996; Intille & Bobick, 1999), discussed in Section 7.

However, a much more important social structure exists in teams. Agents in teams work *together*, as team-member are ideally in *agreement* about their joint goals and plans (Cohen & Levesque, 1991; Levesque, Cohen, & Nunes, 1990; Jennings, 1995; Grosz & Kraus, 1996, 1999; Tambe, 1997; Rich & Sidner, 1997; Lesh, Rich, & Sidner, 1999; Kumar & Cohen, 2000; Kumar et al., 2000). This phenomenon—sometimes called *team coherence* (Kaminka & Tambe, 2000)—holds at different levels in the team. Agents in an atomic subteam work together on the plans selected for the subteam, subteams work together with sibling subteams on higher level joint plans, etc. Individual agents may still choose their own execution, but they do so in service of agreed-upon joint plans. Provided the monitoring agent knows what plans are to be jointly executed by which subteams, and what transitions are to be taken together by which subteams, it can use coherence as a heuristic, preferring hypotheses in which team-members are in agreement about their joint plans, over hypotheses in which they are in disagreement.

For example, suppose that the entire team is known to be executing `Fly-Flight-Plan` (Figure 1-b). Now, a message from one member of the TRANSPORT subteam is observed, indicating that it has begun execution of the `Transport-Ops` plan step. Since this plan step is to be jointly executed by all members of the TRANSPORT subteam (and only them), we can use coherence to prefer the hypothesis that the other subteam members have also initiated execution of `Transport-Ops`. Furthermore, since this plan-step is in service of the `Landing-Zone-Maneuvers` plan, which is to be jointly executed by the TRANSPORT and ESCORT subteams, we can prefer the coherent hypothesis that team-members of ESCORT are executing `Landing-Zone-Maneuvers`. Now, based on their known role, we can now come back down the plan-hierarchy and infer that members of the ESCORT subteam are executing `Escort-Ops`, etc.

This knowledge of the expected relationships, and in particular knowledge of which plans are joint to team-members (i.e., are subject to coherence), is part of the specification of a distributed application—and can thus be provided to an overhearing system by the designer or operator. In fact, it is often readily available, since it is used by the agents themselves in their coordination. For instance, we have earlier discussed the assumption that team-oriented programs are available to the monitoring agent, and that these hold knowledge about what plans in the hierarchy are to be executed by which (sub)teams is encoded in the plan-hierarchy. The team hierarchy contains the knowledge about what subteam/agent is part of another subteam.

Coherence can be a very powerful heuristic. It assumes non-failing cases, where team-members successfully maintain their joint execution of particular plans. Under this assumption, evidence about a decision made by one team-member influences (through coherence), our belief of what its team-mates have decided. And lacking such evidence, coherence prefers hypotheses in which at least the team-members have made joint decisions. For instance, suppose a transition from a team plan is to be taken only by the TRANSPORT team. Under non-failure circumstances, there are only two coherent hypotheses considering this transition: Either all members of TRANSPORT took the transition, or none did. Evidence for one member, supporting one of these hypotheses, can be used to infer the state of the other members.

The significance of this property of coherence is that if the monitoring system can reduce the uncertainty for even one agent, then this reduction will be amplified through the use of the coherence heuristic to apply to the other agents as well. The use of the coherence heuristic can thus lead to a significant boost in monitoring accuracy, since the number of hypotheses underlying any further (probabilistic) disambiguation is cut down dramatically. Section 6.1 provides an in-depth evaluation of the use of coherence and knowledge of roles to select plan recognition hypotheses in Overseer.

The use of coherence significantly increases the time complexity of the computation. At the very least, it requires setting inter-agent links in the array of plan recognizers used by Overseer (Section 3.4), such that these links represent a probabilistic association between plans that are to be executed jointly (in contrast with the temporal and hierarchic decomposition transitions used thus far). For instance, if a specific plan $X$ is to be executed jointly by agents $A$ and $B$, then such a link would be constructed between the variable $X_t^A$ (representing agent $A$'s execution of a plan $X$) and the variable $X_t^B$ (representing agent $B$'s execution of the same plan). In general, there would be $\frac{N*(N-1)}{2}$ such inter-agent links between $N$ agents, for each one of the joint plans (of which there are at most $M$). Thus given $N$ agents, and the array of recognizers $M[]$, where each individual agent's plan-hierarchy is of size $M$, the run-time complexity of an exact-inference algorithm would be at least $O(MN^2)$ and quite likely much worse (since in general there is an exponential number of coherent and non-coherent hypotheses to select from). In the next section (Section 5.1), we describe a highly scalable (in the number of agents) representation for reasoning about coherent hypotheses.

### 4.2 Exploiting Procedures that Maintain Social Structures

A monitoring system can exploit knowledge of the procedures agents use to maintain their social structures to alleviate some of the uncertainty resulting from the scarceness of observations. For instance, if the monitoring system could accurately predict *future observable* behavior of monitored agents, then while it has not observed the predicted behavior, the monitoring system may infer that the agents have not reached the state associated with the predicted behavior. Thus such predictions can be used to eliminate monitoring hypotheses, by setting an individual agent's $\mu_{XY}$ probabilities to reflect a prediction that a message will be transmitted by the agent as its execution of $X$ terminates and it initiates $Y$. For instance, in our own application, the Ariadne information agent is queried for possible threats before each route is followed in the evacuation. It may therefore be possible to predict that before each route is taken by the helicopters, a message will be sent by the Ariadne agent to its teammates; thus while no such message is observed, the Ariadne agent can be inferred to have not yet executed this step. Furthermore, under the assumption of coherence (discussed above), the monitoring system may further infer that all team-members have not yet executed this step, i.e., a new route was not taken by the team. Such inference is obviously dependent on the system's observational capabilities, but we have found it to be useful even under lossy observations by the monitoring system (see Section 6.2).

However, in general, such specific individual predictions can be difficult to make. Team-members are often engaged in joint tasks, which require many agents to tackle a problem together. In these settings, predicting individual communications may be impossible. For instance, consider a distributed search problem in which a target solution is to be found somewhere in the search-space; different areas of the search space are divided amongst the agents, with the understanding that the first to find the target will communicate with the others. It would be difficult to accurately predict which one of the agents will communicate (find the target), since if we could predict that, we could focus all agents' efforts on that area alone. Yet it is easy to predict that at least one agent will find the target and communicate. Similarly, in the evacuation application, it may be difficult to predict which helicopter will reach the civilians first—but it is easy to predict that one of them will, and will then communicate their location.

Indeed, teams utilize *social procedures* or *conventions* (Jennings, 1993) by which team-members maintain their relationships with one another. Removal of the agent independence assumption allows the monitoring system to exploit knowledge of such procedures, by making predictions as to the behavior of team-members in coordinating with one another. For instance, knowledge of the failure-recovery procedures used by a team to recover from coordination failures allows the monitoring system to predict the future behavior of team-members in case of failed execution. Similarly, knowledge of the communication procedures used by the team (as part of its team-members' coordination) allows predicting future observable messages—future interactions between team-members—without necessarily specifying a particular individual agent that will carry them out.

For example, suppose Overseer overhears a message indicating that the flight team has initiated joint execution of `Fly-Flight-Plan` (Figure 1-b). After some time has passed, it is now possible that the team is either still executing `Fly-Flight-Plan`, or it has terminated it already and begun joint execution of `Landing-Zone-Maneuvers`. However, if Overseer knows that at least one team-member will explicitly communicate after terminating `Fly-Flight-Plan` and before initiating `Landing-Zone-Maneuvers`, then while such communications are not observed, the monitoring system can eliminate the possibility that the team is executing the latter, eliminating any uncertainty in this case (only `Fly-Flight-Plan` is possible).

We leave discussion of how technically a social procedure of the form "at least one team-member will communicate when its subteam will take this transition from $X$ to $Y$" can be converted into $\mu_{XY}$ values to the next section, where we present a technique for representing team-wide $\mu$ probabilities in a way that allows efficient reasoning. In the remainder of this section, we address instead how knowledge of such social procedures may be acquired.

Social procedures of communications may be simple per-case rules, or may involve complex algorithms. For instance, Jennings (1993) suggests using heuristic application-dependent rules to determine communication decisions. STEAM (Tambe, 1997) instead uses a decision-theoretic procedure that considers the cost of communication and the cost of miscoordination in the decision to communicate. Other procedures have been proposed as well (e.g., Cohen & Levesque, 1991; Jennings, 1995; Rich & Sidner, 1997). However, regardless of their complexity, a key point is that a monitoring system does not necessarily have to have full knowledge of these procedures in order to exploit them for predictions: it only needs to approximate their outcome, since it can use a combination of techniques to combat plan-recognition ambiguity, rather than relying just on one technique.

The decisions of social procedures can be acquired by learning from previous runs of the system. Although a detailed exploration of appropriate learning mechanisms is outside the scope of this paper, we provide a strict demonstration of the feasibility of learning social procedures by simple rote-learning, which proved effective in generating a useful communications model that significantly reduced the uncertainty in monitoring the evacuation application. This simple mechanism records during execution which plans are explicitly communicated about, and whether they were initiated or terminated. The learned rules are effective immediately, and are stored for future monitoring of the same task.

Figures 4a–d present the results from using of this rote-learning mechanism in four different runs on the same tasks. The X-axis denotes observed communication message-exchanges as the task progresses. Overall, between 22 and 45 exchanges take place in a run, each exchange including between one and a dozen broadcast messages in which agents announce termination or initiation of a plan. The Y-axis shows the number of hypotheses considered by Overseer after seeing each message, without using any probabilistic temporal knowledge. Thus greater uncertainty about which hypothesis is correct would be reflected by higher values on the Y-axis. At the beginning of task execution, all possible plans are considered possible, since we ignore temporal knowledge in this graph. As progress is made on the task, less and less steps remain possible before the end is reached, and so we expect to see a gradual (non-monotonic) decline as we move along the X-axis. A technique that successfully eliminates hypotheses from considerations results in Y-axis values *lower* than those of this baseline execution curve.

In Figure 4, the line marked *No Learning* shows this baseline (i.e., no predictions, and with the learning component turned off). The baseline shows that a relatively high level of ambiguity exists, since the system cannot make any predictions about future states of the agents, other than that they are possible. When the learning technique is applied on-line (i.e., any message seen is immediately used for future predictions), some learned experience is immediately useful, and ambiguity is reduced somewhat (the line marked *On-Line Learning*). However, some exchanges are either encountered late during task execution, or are seen only once. Those cannot be effectively used to reduce the ambiguity of the monitoring system on the first run. However, the third line (*After Learning*) presents the number of hypotheses considered when a fully-learned model is used. Here, the model was learned on run G, then applied without any modifications in the other runs of the system. As can be seen, it shows a significantly reduction in the number of hypotheses considered by Overseer. Further evaluation of the use of communications predictions is presented in Sections 6.1 and 6.2; however, a full exploration of the use of learning for this task is beyond the scope of this paper.

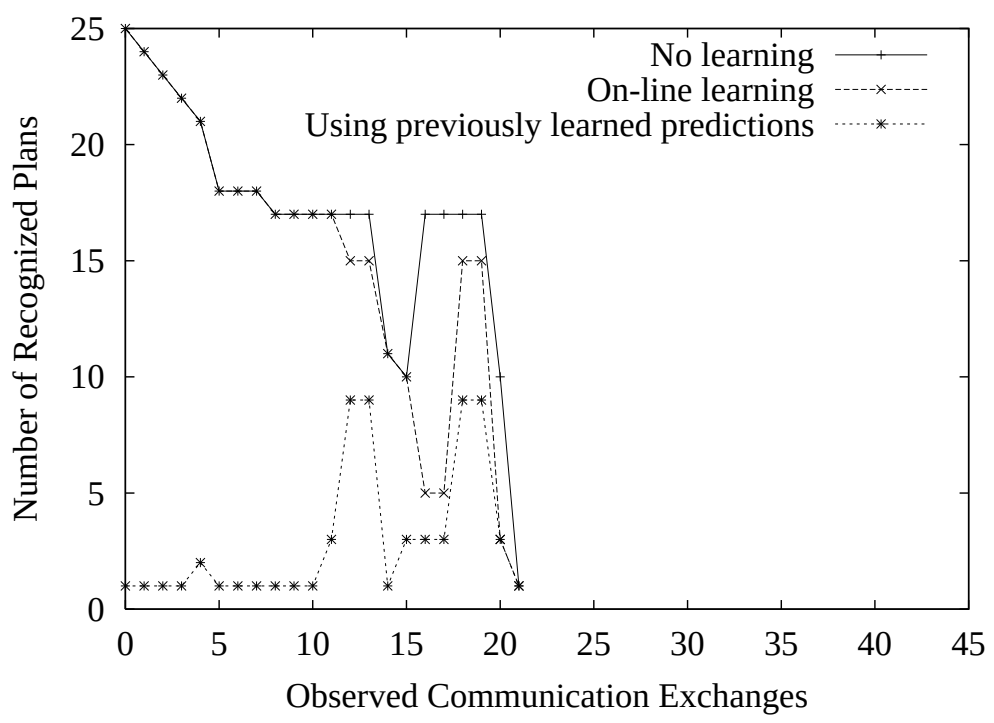

(a) Learning in experiment C

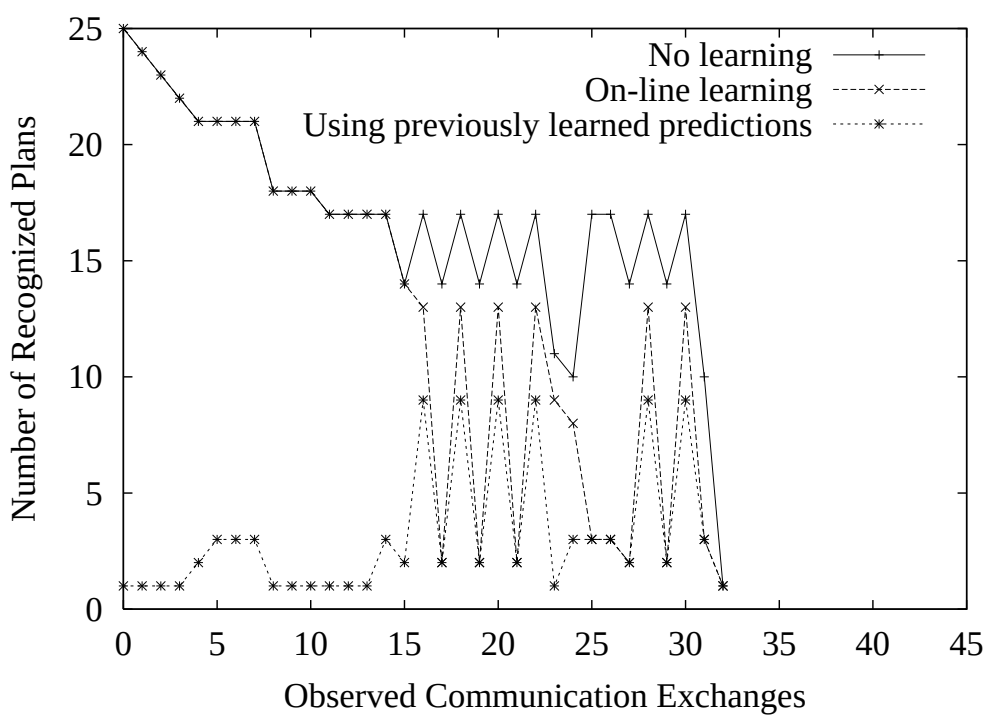

(b) Experiment E

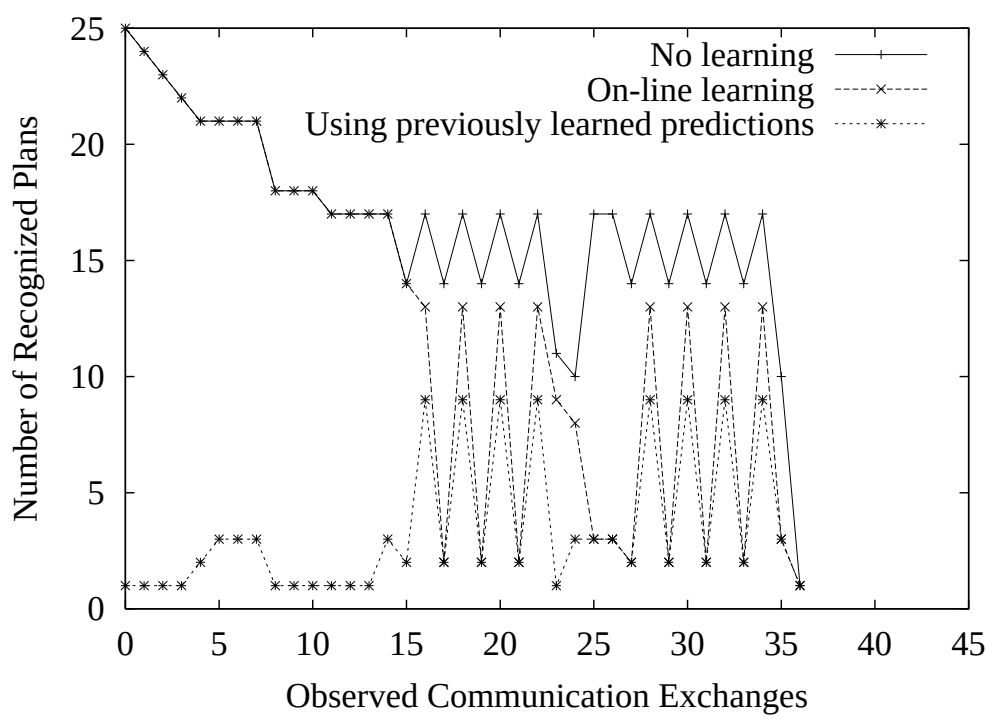

(c) Experiment G

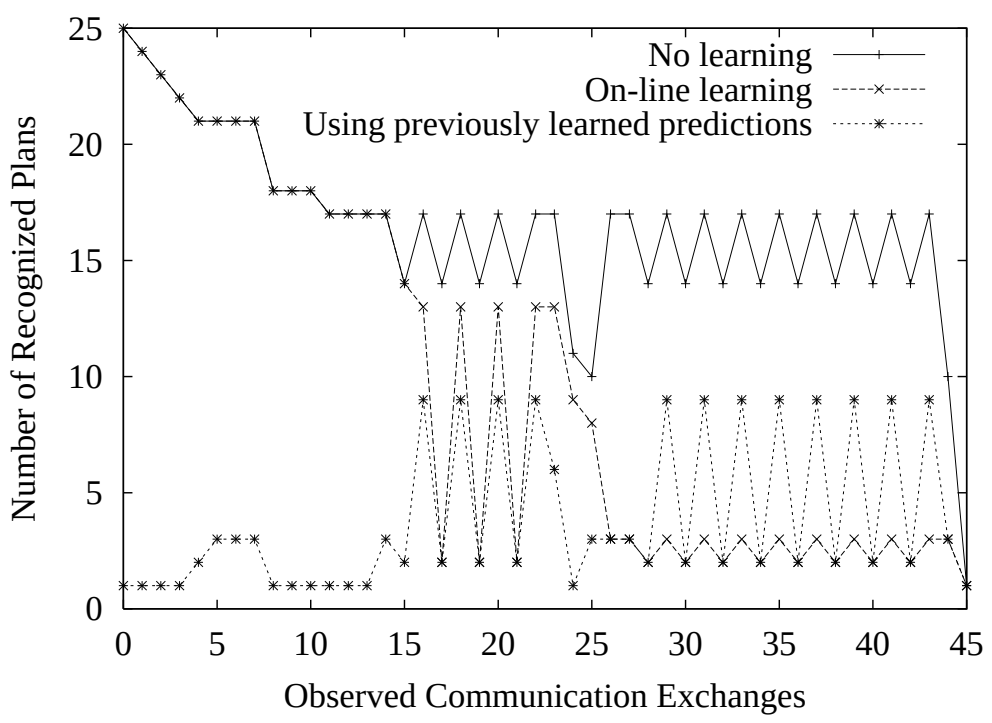

(d) Experiment I

Figure 4: Learning of communication decisions in different experiments.

### 4.3 Discussion

A key characteristic of monitoring by overhearing tasks is the scarcity of observations available to the monitoring system. Fortunately, the observations available to the monitoring system can often be viewed as observations of *multi-agent actions*: The sender of the message not only changes its own state, but often also intends to change the state of the recipients (Cohen & Levesque, 1990). Thus even a single observation can be used as evidence for inferring the state of both sender and receivers. This stands in contrast to previous work, which addressed monitoring of multiple *single-agent* actions.

In monitoring a team, the monitoring system can use knowledge of social structures and procedures to exploit information about the activities of one team-member, in hypothesizing about the activities of another team-member. These techniques are not specific to the representation presented earlier. For instance, an increased belief in one agent's execution of a plan $X$ based on evidence for a teammate's execution of $X$ can be also used by constructing appropriate probabilistic links between nodes representing these beliefs in a large DBN representing the two agents. If we start with the DBN representation as discussed in Section 3.1, we can replicate the single-agent network (containing $M$ plans) for each of the $N$ separate agents. The number of nodes is then $O(M^2N)$, since we represent the plans and transitions for each individual agent. We can also introduce the appropriate inter-agent links to capture the inter-agent dependencies represented by our model of teamwork. However, upon introducing such links, the computational complexity of performing DBN inference explodes to $O(2^{M^2N})$.

Obviously, such "social reasoning" can be computationally expensive, even with the efficient representation described earlier. The next section provides details of an efficient mechanism for reasoning about a team using information about role and coherence, and utilizing communications predictions. Using this mechanism, the techniques described in this section have resulted in an accuracy of up to 97% (84% average across all experiments)—compared to average 4% without the use of social knowledge. Sections 6.1 and 6.2 present a detailed discussion of these results.

## 5. Plan-Recognition for Overhearing

The previous section has outlined socially-attentive monitoring techniques, alleviating the uncertainty in monitoring a team of agents by exploiting knowledge of the social structures and social procedures of the monitored team. It discussed using *coherence* and *role maintenance* to exploit knowledge of the ideal agreement of agents that specific plans be executed together, and that other specific plans are assigned to agents fulfilling their roles. Furthermore, it discussed disambiguation based on predictions of future observable behavior, based on knowledge of the social procedures employed by team-members. These disambiguation heuristics eliminate many (incorrect) hypotheses from being considered. However, reasoning using these techniques can be computationally expensive.

This section presents an efficient algorithm, building on the representation previously presented, which facilitates scalable monitoring by overhearing of large teams. The key idea here is to represent only those hypotheses which the heuristics would have considered valid, eliminating from consideration plans and transitions that would be considered illegal with the heuristics. Relying on the team-hierarchy for bookeeping, all coherent hypotheses are represented using a single recognizer instead of an array of recognizers, offering considerable scalability in team monitoring. However, since the algorithm can no longer represent certain hypotheses, this scalability comes at the expense of expressivity. We discuss the scalable representation and the trade-off it offers below.

### 5.1 Efficient Reasoning with Team Coherence

Coherence is a very strong constraint, since for a team of agents there are only a linear number ($O(M)$ where $M$ is the size of the plan-hierarchy) of coherent hypotheses, but an exponential number of incoherent hypotheses ($O(M^N)$ where $N$ is the number of agents; the proof is in Appendix A). We can exploit this property by designing monitoring algorithms that reason only about the linear number of coherent hypotheses, and therefore offer better scalability as the number of agents increases. Such algorithms may not be able to reason about incoherent hypotheses, and are therefore less expressive. However, Section 6 demonstrates that the level of accuracy even with such limited expressiveness is sufficient for our purposes. Furthermore, algorithms that reason only about coherent hypotheses may still be able to detect incoherent hypotheses, representing a failure state in which two or more team-members are in disagreement with each other.

We begin by presenting the YOYO* algorithm, an efficient technique for reasoning about coherent hypotheses (Algorithm 5). YOYO* replaces the array-based algorithm described earlier (Algorithm 4). Similarly to it, YOYO* is called every time tick. If no message is observed, the state of the entire team is propagated forward in time. Otherwise, all observed messages are collected together and used as evidence for the (different) plans implied by these messages.

YOYO*'s key novelty is that it relies on a *single* plan-hierarchy that is used to represent all team-members together (regardless of their number), instead of an array of such structures. In other words, each variable $X_t$ represents Overseer's belief that *all agents* in the teams associated with $X$ (as described in the team-oriented program) are executing the plan $X$ at time $t$. Thus YOYO* makes extensive use of the information associating plans and transitions in $M$ with teams and subteams in $H$, the team-hierarchy. The team hierarchy plays a critical bookeeping role in this respect, since it maintains the knowledge critical for correctly applying coherence in the single recognizer.

This key distinction between YOYO* and the array-based approach causes a subtle, but critical, difference in the way probabilities are propagated along transitions. In a plan-hierarchy $M$ of an individual agent, part of an array of such models, each outgoing transition represented a hierarchical decomposition or temporal step that the agent is allowed to take. Alternative outgoing transitions therefore represent alternative paths of execution available to the agent. On the other hand, in a plan-hierarchy $M$ used by YOYO*, alternative outgoing transitions tagged by different subteams (that are not ancestors of one another) represent not a decision point for the agent, but alternative paths of execution as decided by the agents' roles and team-memberships.

This creates a critical difference in how the values of $\pi_{XY}$ and $\mu_{XY}$ are to be interpreted. Where previously (in Section 3) the value of $\pi_{xy}$ referred to the probability that a specific agent will take a transition $X \rightarrow Y$ (given that it has terminated execution of $X$), in YOYO* it refers to the probability that an entire team will take the transition *together.*

**Algorithm 5** YOYO*(**plan-hierarchy** $M$, **team-hierarchy** $H$, **beliefs** $b$)

**if** no new messages are observed **then**
  TEAM-PROPAGATE-FORWARD($b$, $M$)
**else**
  Initialize distributions $b', b_{t+1} \leftarrow 0$ for all plans $U \in M$. ; Initialize $I, E$ to be empty sets.
  **for all** Messages $m_i$ **do**
    $I \leftarrow I \cup \{X \mid X \in M,\ m_i$ is a an initiation message, $X$ consistent with $m_i\}$
    $E \leftarrow E \cup \{Y \mid Y \in M,\ m_i$ is a termination message, $Y$ consistent with $m_i\}$

  **for all** plans $X \in I$ **do**
    $T \leftarrow team_{msg}(X)$ $\{T$ is the agent sending the message initiating $X\}$
    **for all** plans $W \in M$ that precede $X$, where the transition $W \rightarrow X$ is allowed for $T$ **do**
      $b'(X, \neg block) \leftarrow b'(X, \neg block) + b_t(W, block)\mu_{wx}\pi_{wx}$
  **for all** plans $X \in E$ **do**
    $T \leftarrow team_m sg(X)$ $\{T$ is the agent sending the message terminating $X\}$
    **for all** plans $Y \in M, Y \notin I$ that succeed $X$, where the transition $X \rightarrow Y$ is allowed for $T$ **do**
      $b'(Y, \neg block) \leftarrow b'(X, \neg block) + b_t(X, block)\mu_{xy}\pi_{xy}$
  Normalize distribution $b'$ taking teams into account
  **for all** plans $X$ where $b'(X, \neg block) > 0$ **do**
    $b_{t+1}(X, \neg block) \leftarrow b'(X, \neg block)$
    TEAM-PROPAGATE-DOWN($X, b'(X, \neg block), b, M$)
    $T \leftarrow team(X)$
    $P \leftarrow X$
    **while** $parent(P) \neq null$ **do**
      $b_{t+1}(parent(P), \neg block) \leftarrow b_{t+1}(P, \neg block)$
      **if** $team(parent(P)) = parent_{team}(T)$ **then**
        SCALE($parent(P), T, P, b$)
        $T = parent_{team}(T)$
      $P \leftarrow parent(P)$

YOYO* is unable to represent hypotheses in which some team-members take one transition, and others do not—unless these two different groups of members form different subteams that are represented in the team-hierarchy, and the different transitions are tagged as being allowed for the different subteams.

The value of $\mu_{XY}$ is also interpreted differently, in a very critical way. Where in the previous sections it was taken to represent the probability that a specific individual will communicate when a transition $X \rightarrow Y$ is taken, in YOYO* its value represents instead the probability that *one or more team-members will communicate* when the transition is taken by the team. Thus it no longer refers to individual agents, but to a (sub-)team. In this way, YOYO* solves the issue of how to represent predictions of the type "at least one team-member will communicate when this step is reached", discussed previously.

For example, suppose YOYO* sets the belief that the team is executing the `Landing-Zone-Maneuvers` plan-step to some probability $p$. `Landing-Zone-Maneuvers`, in YOYO*, has two (first) children: `Escort-Ops` and `Transport-Ops`, to be executed by members of th ESCORT and TRANSPORT subteams, respectively. Unlike in the individual agent case, the probability $p$ should not be divided among these two children, but should be duplicated to them: A belief that the entire team is executing `Landing-Zone-Maneuvers` implies an equally-likely belief that the TRANSPORT subteam is executing `Transport-Ops`, and that the ESCORT subteam is executing `Escort-Ops`. We explain YOYO*'s operation in detail below:

**No message is observed (lines 1–2).** Since no observations are available, the state of the entire team is jointly propagated forward in time by calling TEAM-PROPAGATE-FORWARD (Algorithm 7, Appendix A). This is a slightly modified version of the PROPAGATE-FORWARD (Algorithm 3) that takes different subteams into account in propagating beliefs: Given some total outgoing probability (either to a sibling or child transition), if the outgoing transitions are to be taken by different teams where one team is not an ancestor of another (such as the TRANSPORT and ESCORT sub-teams), the same total probability would be used for each transition, instead of splitting the outgoing probability between the transitions. Appropriately, TEAM-PROPAGATE-FORWARD relies on a modified version of the PROPAGATE-DOWN algorithm (Algorithm 2), called TEAM-PROPAGATE-DOWN (Algorithm 6, Appendix A). This latter algorithm is also used in the incorporation of evidence (lines 3–27). The run-time complexity of the propagation process is $O(M)$.

**One or more messages are observed (lines 3–7).** If one or more messages are observed (since YOYO* is a single algorithm monitoring multiple potential message senders, more than one message may be observed at once), YOYO* begins to incorporate these observations into the maintained beliefs about the team. This process is somewhat similar to the INCORPORATE-EVIDENCE algorithm, described earlier (Algorithm 1), but takes into account multiple observations (since all $N$ agents may have sent a message). Multiple messages (from different agents) may all refer to the same plan, but YOYO* must not incorporate evidence for them multiple times.

The simple loop (lines 5–7) builds the set $I$ (of initialized plans) and $E$ (of terminated plans) by going over all incoming messages that have arrived at time $t$. The run-time complexity of this process (in the worst case) is $O(N)$. Here, YOYO* does better than the array approach, since multiple messages always cause multiple updates in the array, but in YOYO*, multiple messages may all refer to a single plan, thus triggering a single update.

**Incorporating evidence about initiated and terminated plans (lines 8–15).** For each one of these plans $X$ in $I$ (line 8), YOYO* now sets the new belief $b'$, weighted by any prior belief in $X$'s initiation (lines 10–11), similarly to how this is done in the INCORPORATE-EVIDENCE algorithm (Algorithm 1), but taking into account the team implied by the sender of the processed message (line 9). This is done by a lookup into $M$ using the sender $T$ ($team_{msg}(m_i)$): Only transitions in $M$ that $T$ is allowed to take are followed. By definition, any transition that is allowed to be taken by a super-team of $T$ is allowed for $T$. A similar process is then done with any termination messages (lines 12–15), but of course looking at possible successors of any plans consistent with the messages. However, since we do not want to cause updates in both line 11 and line 15 in cases where a termination message and an initiation message refer to the same transition, the loop over the plans $Y$ (line 14) skips any plans which have already been addressed in the previous step. Overall, the run-time complexity of this process is $O(M)$.

**Normalizing the temporary distribution $b'$ (line 16).** The temporary distribution $b'$ resulting from the processing of initiation and termination messages is normalized, in a similar fashion to the analogous step in algorithm 1. However, the process must take into account not only the plan-hierarchy in question, but also the team-hierarchy. Unlike a typical normalization procedure, evidence for two different plans, selected by two different teams, may not necessarily compete with each other, and therefore may not necessarily require normalization. For instance, if two messages are observed, one implying that team $A$ has initiated execution of plan $P$, and another implying that team $B$ has initiated execution of plan $Q$, then if $P$ and $Q$ are both children of a joint parent $J$ (executed jointly by the two subteams $A, B$), then the same normalized likelihood (1.0) should be assigned to $P$ and $Q$ (and $J$—but this will be assigned to it by the propagation steps described below). The run-time complexity of this process is $O(M)$.

**Propagating the evidence up and down $M$ (lines 17–27).** First, the beliefs are set for each plan implied by the observations, and its children (lines 18–19). Then, the team $T$ that is to execute this plan is determined by a lookup into $M$ using $team(X)$ (line 20). Now YOYO* begins to propagate the evidence up to the plan's parents (lines 21–26). Any belief in the child plan is propagated and added to the belief in its parent (line 23). However, if the parent plan ($parent(P)$) is to be executed by a super-team of the current team $T$, then any change to its probability must be propagated to its other children, that are to be executed by other (subteams). Thus the upward propagation is alternated with downward propagation along hierarchical decomposition transitions[1]. This downward step is executed whenever the team that is responsible for joint execution of the parent plan is no longer the current subteam being considered ($T$), but its parent team in the team-hierarchy $H$, given by $parent_{team}(T)$ (lines 24–26). When this condition is satisfied, any change in the beliefs about the parent plan must be propagated down to any children it has that are to be executed by other subteams. This is done via the SCALE algorithm (Algorithm 8, Appendix A).

1. This alternating upward-downward propagation is the origin for YOYO*'s name.

The downward propagation (line 25) implements a subtle but critical step: It re-aligns any beliefs YOYO* maintains about subteams other than those implied by the message so that these beliefs are made coherent with existing evidence. The Scale procedure, which re-distributes the new state probability of a parent among its children, such that each child gets scaled based on its relative weight in the parent. The end result is that the state probabilities of the children are made to sum up to the state probability of the parent. The process is recursive, but never re-visits a subtree, since it is only carried out for hierarchical-decomposition transitions that were not previously updated.

Once this downward propagation is done, YOYO* updates the current team to be its parent in the team-hierarchy, in line 26. Note that the call to $parent_{team}$ reflects a lookup in the team-hierarchy $H$, rather than the plan-hierarchy $M$. Finally, regardless of whether downward propagation took place, the temporary variable $P$ is updated to climb up the hierarchical decomposition in $M$ (line 27).

Each iteration through the loop begun on line 17 is $O(M + H)$ since in the worst case both the plan-hierarchy and team-hierarchy are traversed. However, this loop many repeat (in the worst case) for each of the plans in the plan-hierarchy, and thus overall, the run-time complexity of this process is $O(M(M + H)) = O(M^2 + MH)$.

**An example run of YOYO*.** The following example illustrates YOYO*'s inference upon an observation of a message. Suppose a single member of the TRANSPORT subteam communicates that it is initiating the `Transport-Ops` plan. Upon observing this message, YOYO* looks up the sender, to determine what transitions can be taken by it (line 8). It then proceeds to determine the new beliefs in team $T$'s execution of the `Transport-Ops` plan (lines 9–10, then 16), and incorporates these new beliefs to reflect a much increased belief that the TRANSPORT subteam is executing `Transport-Ops` and its children (lines 18–19). Since this plan's parent, `Landing-Zone-Maneuvers`, is not null, YOYO* enters the loop in lines 22–27. First, it increases the belief in the execution of the parent (line 23). Then, it checks the condition on line 24: Indeed, the team that is to execute `Landing-Zone-Maneuvers` is TEAM-FLY-OUT, the parent of the TRANSPORT subteam (i.e., `Landing-Zone-Maneuvers` is to be executed jointly by the TRANSPORT and ESCORT subteams). YOYO* therefore calls the Scale procedure (line 25) to re-adjust `Landing-Zone-Maneuvers'` other children subtrees. `Landing-Zone-Maneuvers` has two hierarchical-decomposition children: `Transport-Ops` (which YOYO* has already updated) which is to be executed by the TRANSPORT subteam, and `Escort-Ops`, which is to be executed by the ESCORT subteam. Scale climbs *down* from `Landing-Zone-Maneuvers` to `Escort-Ops`, increasing YOYO*'s beliefs that the ESCORT team is executing the `Escort-Ops` plan. This process re-aligns any prior beliefs YOYO* had about the likelihood that `Escort-Ops` was being executed with current evidence, in effect updating beliefs about the plans executed by the ESCORT subteam, based on a single observation made of a member of the TRANSPORT team. The process now repeats this loop until the entire set of beliefs is updated and aligned with respect to the observed message.

### 5.2 Scalability in the Number of Agents

YOYO* offers significant computational advantages when compared to the individual representation (array) approach. YOYO* requires only a *single*, fully-expanded plan-hierarchy to represent the entire team. This hierarchy is *a union* of all the individual agent plan-hierarchies, containing all transitions and plans, tagged by the subteams that are allowed to execute them. In addition YOYO* uses a single copy of the team hierarchy. Suppose $M$ is the size of the plan-hierarchy, $H$ is the size of the team-hierarchy, and $N$ the number of agents in the team. When agents are added to the monitored team, the team hierarchy grows by one new node that represents the new agent, and is connected to the appropriate sub-team in the team hierarchy. YOYO*'s space complexity is therefore $O(M + H)$. Since $H$ grows with $N$, we could write it $O(M + N)$ (compare to the array approach: $O(MN)$, Algorithm 4).

To analyze YOYO*'s run-time complexity, we have to consider the behavior of Algorithm 5 separately in cases where no communications are observed, and in cases where at least one message is observed. If no messages are observed, then an update takes the form of a single call to Team-Propagate-Forward (Algorithm 7), an $O(M)$ process. This is clearly a best-case scenario for YOYO*. If one agent communicates, then YOYO* would have to go through $M$ and $H$ in its upward-downward propagation process only once, thus $O(M + H) = O(M + N)$.

The worst case scenario for YOYO* occurs if all agents send messages, and each one of these $N$ messages refers to a different plan (messages about the same plans would be merged in lines 5–7). In this case, there would be up to $M$ different plans for which evidence exists, and each one of them would require a separate update through lines 17–27. Thus YOYO*'s run-time complexity in this case is

$$O(N + M + M + M(M + H)) = O(N + M^2 + MH) = O(N + M^2 + MN)$$

Clearly, this worst-case cannot be continuously sustained by a monitored team, since agents cannot continuously communicate about their state. We thus believe that the average case in real-world domains with *many* agents would be much closer to the $O(M + N)$ case presented earlier (see Section 6.4 for empiric evaluation). In any case, YOYO*'s complexity compares favorably with a procedure reasoning about coherent hypotheses using an array of recognizers, an $O(MN^2)$ process (at least), even if only one agent communicates (Section 4.1).

### 5.3 Discussion

YOYO* explicitly represents a team as a single coherent entity. Its space and run-time requirements are preferable to the array based approach when the number of agents grow, and it considerably simplifies reasoning about coherence and communications predictions. On the other hand, YOYO* sacrifices the capability to represent failing team activities (incoherent hypotheses), where one team-member is executing one team-plan while its teammate is executing another. This does *not* at all mean that individual actions taken by agents are somehow locked together in synchronous execution, or that individual agents must all execute the same individual action at the same time. For instance, two team-members $A$, $B$ that are each executing a completely different path of execution at the same time (i.e., plan steps $A_1, ..., A_k$ and $B_1, ..., B_l$) can be easily represented by a plan hierarchy that includes an overall joint plan $J$, having two first hierarchical decomposition children, $A_1$ and $B_1$, to be selected by $A$ and $B$, respectively. $A_1$ would have an outgoing temporal transition to $A_2$, etc. and similarly $B_1$ would have an outgoing temporal transition to $B_2$, etc. Since $J$ is to be executed by the two team-members jointly, any initial evidence for any one of the agent executing any of its individual plans would be used by YOYO* as evidence for the other team-member having begun its own parallel execution of its own individual execution path. Further evidence about one agent executing its own individual actions would only increase the likelihood that the other agent is continuing its own execution, at its own pace. However, it would be impossible for YOYO* to correctly represent a monitoring hypothesis in which $A$ is executing some child of $J$, $A_i$, while $B$ is executing some plan that is not $J$, nor a child of $J$. Given the results of the evaluation we conducted (Section 6), which demonstrated the importance of coherence in accurate visualization, the tradeoff of expressivity vs. scalability is justified: Overseer's accuracy was much improved due to the use of coherence.

Although YOYO* sacrifices the capability to reason about certain failure (incoherence) hypotheses, it is still capable of supporting failure-detection, an important secondary goal of visualization. In earlier work, we have shown the merits of coherence in service of detecting disagreements in a team, in particular demonstrating that coherent monitoring leads to sound centralized disagreement detection, and may lead to sound and complete disagreement detection under specific circumstances (Kaminka & Tambe, 2000). As YOYO* is in fact a very efficient way to reason about coherent hypotheses, it provides a good basis for providing sound disagreement detection results.

A concern about the generality of the technique may be raised based on YOYO*'s reliance on the team-hierarchy. However, we believe it is reasonable to expect that large, complex, real-world multi-agent systems of the type targeted by this paper would have an organizational hierarchy of some sort associated with them (see, for instance, Tidhar, 1993b). Human organizations certainly demonstrate the emergence of such hierarchies, especially as the organizations grow larger (e.g., big corporations, government organizations) or tackle mission-critical tasks (e.g., military organizations). In addition, team-hierarchies for computational agents are critical for planning, for maintaining network and system security, etc. Thus we believe our use of a team-hierarchy is not a weakness in our approach, as organizational structures will become as wide-spread in computational multi-agent systems as they already are in human multi-agent systems. Indeed, it may be possible to gradually learn a team-hierarchy for a given coordinated team for the purpose of monitoring; however, discussion of this possibility is outside the scope of this paper.

Indeed, using a team-hierarchy, we can apply our assumption of coherence to other representations and algorithms as well. For instance, if we start with the DBN representation of the team from Section 4.3, we can unify the multiple random variables used to represent the separate agents into a single random variable for an overall team/subteam. As in YOYO*, the size of the representation grows with the size of the plan hierarchy, and not the number of agents. Thus, the number of nodes will be the same as for the single-agent case, $O(M^2)$, as discussed in Section 3.1. However, again, the complexity of inference in answering plan-recognition queries will still be exponential in the number of nodes, $O(2^{M^2})$.

## 6. Evaluation

This section presents a detailed evaluation of the different contributions contained within Overseer. We begin by exploring the relative contribution of each technique to the success of Overseer as a whole (Section 6.1). We then focus on evaluating Overseer's use of communications predictions with respect to lossless and lossy observations (Section 6.2). We then present a comparison of Overseer's performance with that of human experts and non-experts (Section 6.3). Finally, we empirically evaluate YOYO*'s scalability in our application domain (Section 6.4).

### 6.1 Accuracy Evaluation

The first part of the evaluation tests the contribution of the different techniques in Overseer to the successful recognition of the correct state of the team-members. Figure 5 compares the average accuracy for a sample of our actual runs, marked A through J (X-axis). In each such 10–20-minute run, the team executed its task completely. At different points during the execution, the *actual* state of the system was compared to the state *predicted* by Overseer, where the prediction was taken to be the current most-likely hypothesis. Each run had 22–45 such comparisons (data-points). The percentage of correct monitoring hypotheses for each run across those comparisons is given in the 0-1 (0-100%) range, on the Y-axis.

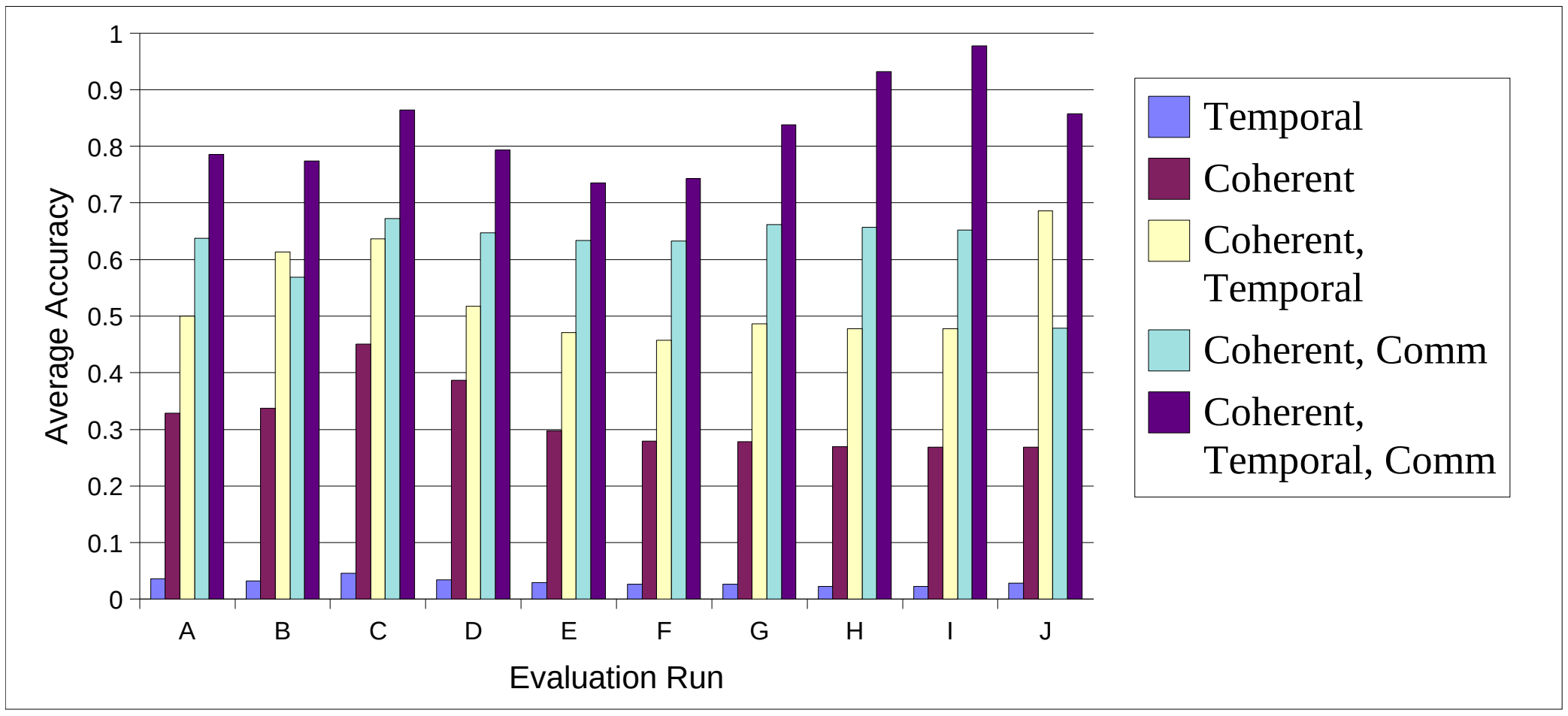

Figure 5: Percent accuracy in sample runs.

The accuracy when using the individual models with no coherence (as in Section 3) is presented in the leftmost bar (marked *Temporal*) in each group (Figure 5), and is clearly very low. This approach is a straightforward attempt at monitoring multiple agents by monitoring each individual, without considering the interactions between them, as described in Section 3. The next bar presents the monitoring accuracy when only coherence is used to rule out hypotheses (Section 5.1), with ties broken randomly. The next bar to the right (*Coherent*, *Temporal*) presents the results of combining both coherence and the probabilistic temporal model (Sections 3 and 5.1). Then, the bar marked (*Coherent, Comm*) shows the effects of combining the use of coherence with the use of predictions based on knowledge of the communication procedures used by the team (Section 4.2). Here, the communications predictions were used to restrict the set of coherent hypotheses considered, with ties broken randomly. The remaining bar (*Coherence, Temporal, Comm*) presents the monitoring accuracy in each run using the combination of all techniques.

The results presented in Figure 5 demonstrate the effectiveness of the socially-attentive monitoring techniques we presented. First, the results show that the coherence heuristic brings the accuracy up by 15–30% without using any probabilistic reasoning. This boost in performance is a particularly interesting result, because of the relation between the coherence technique and previous techniques explored in the literature (Tambe, 1996; Intille & Bobick, 1999). Previous work has successfully used the relationships between agents to increase the accuracy of monitoring. The boost in Overseer's accuracy based on the use of role and teamwork relationships confirms the results from previous investigations. However, the results also demonstrate that the technique is not sufficient in this domain.

Overseer adds a number of novel techniques not addressed in previous work. The first such technique combines coherence with a temporal model of plan-duration (*Coherent, Temporal*), and it results in significant increases to the accuracy, because the probabilistic temporal information now allows Overseer to better handle the lack of observations. A possible alternative, which we explore in this evaluation, is to rely instead on the communications predictions to rule out hypotheses about future states that may or may not have been reached (*Coherent, Comm*). It is therefore interesting to compare the performance of these two techniques by comparing the (*Coherent, Temporal*) and (*Coherent, Comm*) bars.

In almost all runs the average accuracy when using coherence and communications predictions is significantly higher than when using coherence and the temporal model. This is despite the fact that the more effective coherence technique uses arbitrary (random) selection among the available hypotheses: The reason for this is that in many cases the communication predictions are powerful enough to rule out all hypotheses but one or two, significantly decreasing the uncertainty of the agents' plan-horizons. Thus even a random selection stands a better chance than a more informed (by a temporal model) selection among many more (10–20) hypotheses.

However, runs J and B show a reversal of this trend compared to the other runs. Figures 6a–b show the accumulative number of errors as task execution progresses during run I (Figure 6-a) and during run J (Figure 6-b). An error is defined as a failure to choose the correct hypothesis as the most likely one (i.e., the most likely hypothesis does not reflect the true state of the agent/team). Each message exchange corresponds to one to a dozen messages communicated by the agents, establishing or terminating a plan. In the two figures, a lower slope means better performance (less errors). The line marked *Coherent* shows the accumulative number of errors if only coherence is used to select the correct hypothesis—most such choices turn out to be erroneous since a random choice is made among the competing hypotheses. The line marked *Coherent, Temporal* shows the results using both coherence and the temporal model to choose the most likely hypothesis. Similarly, the line marked *Coherent, Comm* shows the results using both coherence and the communications predictions. Finally, the remaining line displays the results of using the combined technique, using coherence, the temporal model, and the communications predictions.

In Figure 6-a, we see that the two techniques (*Coherent, Temporal* and *Coherent, Comm*) have almost equal slopes and result in almost equal number of errors at the end of run I, though from Figure 5 we know that due to the alleviated uncertainty, the use of communications predictions leads to overall higher probability of success (i.e., the *Coherent, Comm* technique results in fewer alternative hypotheses, and thus has a better chance of being correct). However, in Figure 6-b we see that in run J the situation has changed dramatically. First, we see that the two lines are no longer similar. The line marked *Coherent, Comm* has greater slope than in run I, indicating that the communications predictions are not able to reduce the uncertainty, resulting in lower average accuracy. Second, we see that the temporal model results in many less errors, as evidenced by the much slower-rising slope of the line marked *Coherent, Temporal*. Thus in this case, the actual duration of plans matched the temporal model more accurately than in other runs.

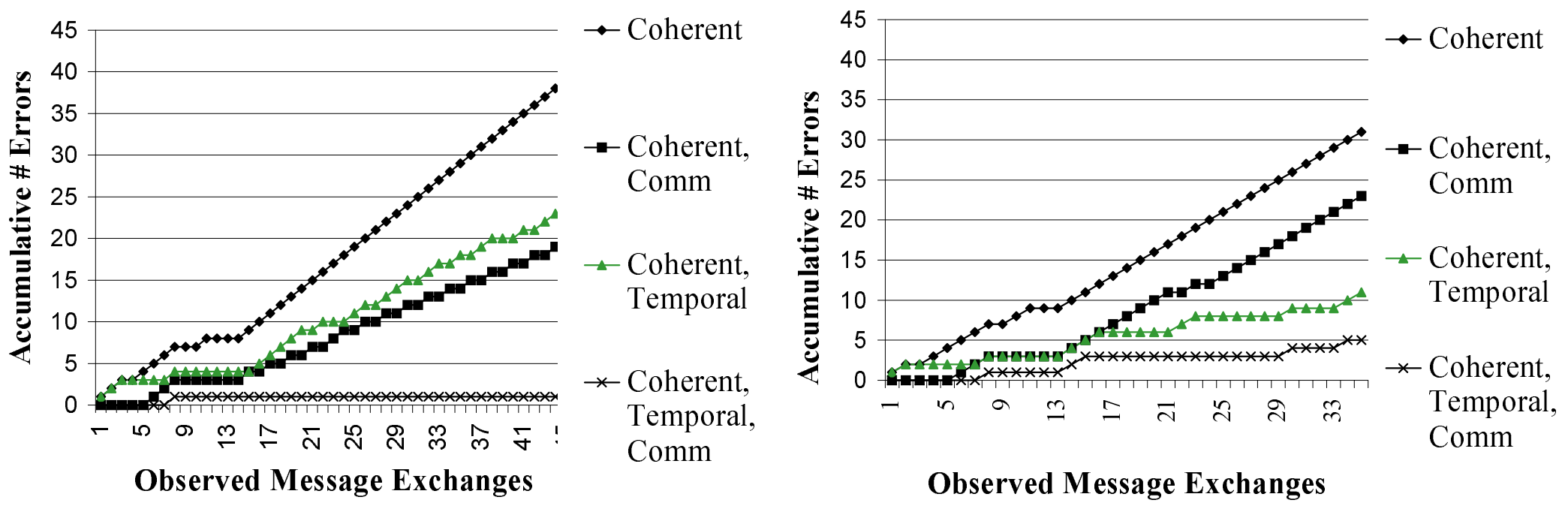

(a) Run I (b) Run J

Figure 6: Accumulative number of errors in runs I and J.

In trying to understand this difference between runs J, B and the other runs of the system, we discovered that runs J and B involved relatively more failures on the part of team-members, including agents crashing or not responding at all. The communications predictions, however, were learned based on successful runs—and thus did not correctly predict the communication messages that would result as the team detected and recovered from the failures. Thus the uncertainty was not alleviated, and the arbitrary selection was made among relatively many hypotheses. This explains the relatively lower accuracy of the (*Coherent, Comm)* technique in run J and B. This clearly shows a limitation of the simple learning approach we took, and we intend to address it in future work. However, there are other factors that influence the accuracy of the communication models, since this lower accuracy did not occur in other runs where failures have occurred.

The results of the *Coherent, Temporal* technique vary as well. We have been able to determine that failures cause a relative increase in the relative accuracy of the *Coherent, Temporal* technique. However, variance in the results is due to additional factors. In run C, for instance, this technique results in relatively higher accuracy, but no failure has occurred. Certainly, the mission specifications themselves differ between runs, machine loads cause the mission execution to run slower or faster, etc. The great variance in the temporal behavior of the system was the principal reason for our using the communication prediction. This variance is obvious in the graphs.

In summary, despite the variance in the results of the *Coherent, Temporal* technique (due to variance in the temporal behavior of the system and the simplicity of the temporal model), and the possible sensitivity of the *Coherent, Comm* technique to learned predictions, it is clear that the two techniques work well in combination, building on the coherence heuristic, and compensating for each other's weaknesses. In all runs, the combined technique *Coherent, Temporal*, *Comm* was superior to either technique alone. Its performance varied between 72% accuracy (Run E) to 97% (Run I). The average accuracy across all runs of this all-combination technique was 84%, resulting in very significant increases in accuracy compared to the initial solution with which we began our investigation (less than 4%), and to human novice performance (see Section 6.3). Thus the communications predictions need not be perfect, and the temporal knowledge need not be precise, in order to be useful.

## 6.2 Evaluating the Use of Communications Predictions

One key question about the use of the communications predictions is their sensitivity to loss of observations. The efficacy of the technique (see Figure 5) stems from its capability to make inferences based on an expected *future* observation. The predictions used in the previous section assumed no observation loss, i.e., if a prediction stated that a particular message was to be observed, than the probability assigned to this prediction was 1.0. But in settings involving lossy observation streams, such inference will prove incorrect, as Overseer will "wait" for the observation and will therefore not correctly monitor the actual state of team-members.

To evaluate the predictions' sensitivity to observation loss, we chose three of the experimental runs, E, I, and J, which represent the extreme performance results of Overseer: Run E had the lowest accuracy (72%), Run I had the highest (97%), and run J showed an interesting reverse in relative performance of the *Coherent, Temporal* and *Coherent, Comm* (see Figure 5). For each of these runs, we simulated observation loss at a rate of 10%, repeating each trial three times with different random seeds. In other words, we ran a total of 9 trials, in which a random 10% of the messages to be observed by Overseer were not observable to Overseer (though they still reached the evacuation team-members—team-performance was identical to the original settings). We then set the predictions to appropriately use 90%–10% settings: each expected message was predicted to appear with 0.9 probability (as opposed to 1.0 probability originally).

The results of these experiments are presented in Figure 7. For each of the three different runs, two bars are presented. The left (shaded) bar shows the original results as presented in the previous section (i.e., with no observation loss, and no treatment of possible loss in the predictions). The right bar shows the average accuracy achieved by Overseer on the three trials (for each run) in which 10% of the observations were not observable to Overseer. The error-bars on the right bar mark the minimum and maximum accuracy values achieved in the three trials for each run. Run I's error-bars are unseen since all three trials resulted in the same accuracy.

There are a number of promising conclusions that can be drawn from these results. First, in both runs E and I, Overseer's average accuracy dropped by less than 8%, i.e., the performance of Overseer dropped by less than the level of loss introduced. Indeed, in run E, in which the original performance was the poorest, there was almost no change in performance. Performance in run J did drop by slightly more than 10%, and that can be at least partially explained by run J's previously discussed failures to exploit the communications predictions. Thus one promising conclusion to be drawn from these results is that Overseer's performance can degrade gracefully, at a rate comparable to the rate of degradation to Overseer's input.

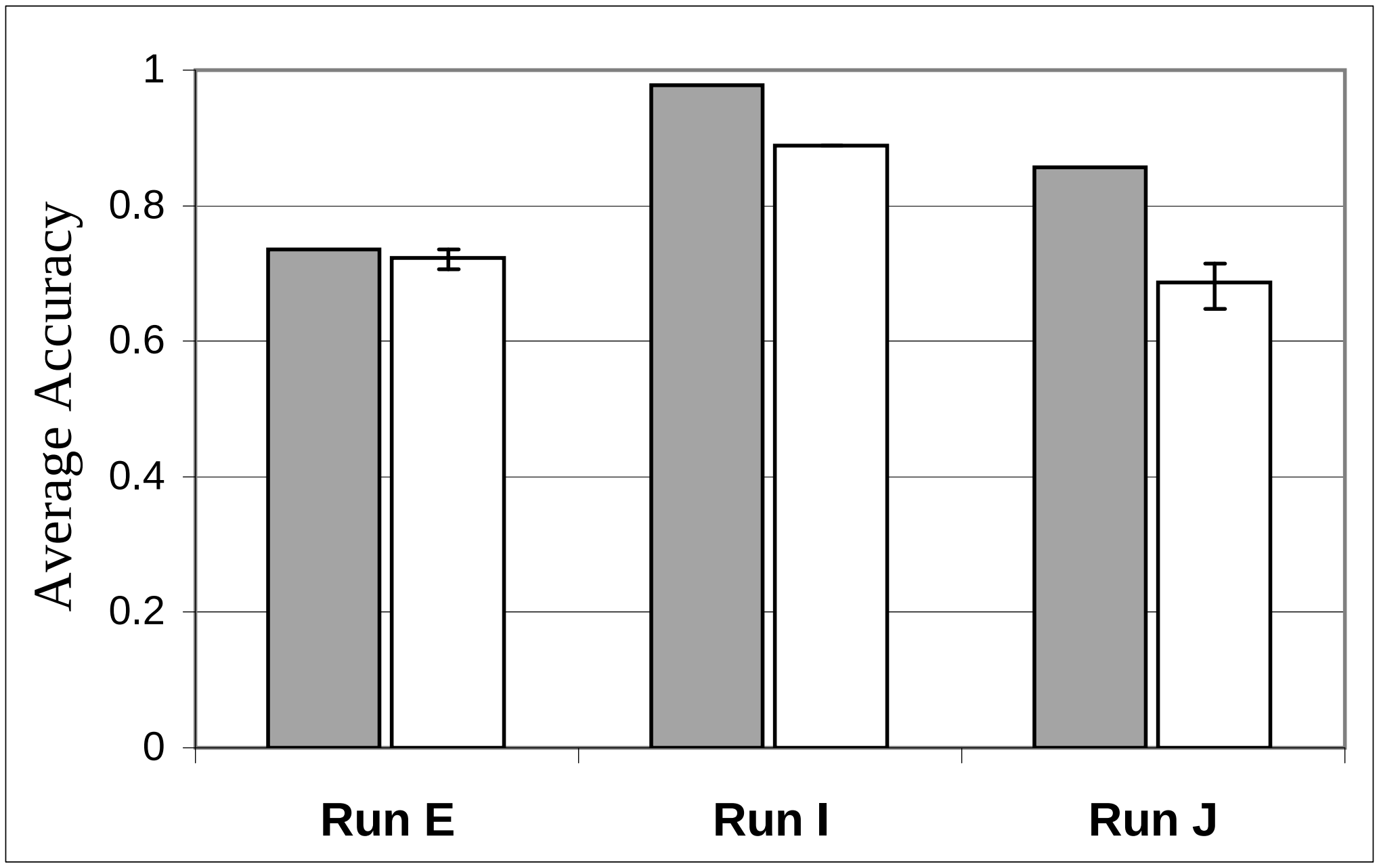

Figure 7: Comparison of average accuracy results with 0% and 10% observation losses.

A second conclusion is that Overseer's performance under observation-loss settings is fairly invariant. Again, both run E and I, which can be considered normative, show very little (if any) variance from one trial to the next, despite the change in the selection of observations to be made unobserved from one trial to the next. Even run J, which is not a representative of the normative runs, shows little variance with respect to its average accuracy under observation loss. This result suggests that while there may be a drop in performance with observation loss (as expected), Overseer performs consistently under varying lossy settings.

### 6.3 Overseer and Human Monitoring by Overhearing

Another important facet to the evaluation of Overseer examines its performance in comparison to that of novice and expert monitors of the evacuation application. This evaluation sheds some light on the difficulty of the monitoring task, and demonstrates that Overseer's performance is comparable (sometimes higher, sometimes lower) to human expert performance, and significantly better than that of novices.

To conduct this evaluation, we examined the same three runs representatives of Overseer's bounds on performance discussed above (runs E, I, and J). The first author of this paper served as an expert monitor, having as much experience in overhearing in the evacua-tion application as possible (and specifically in the actual test runs E, I and J)[2]. We established a group of novice monitors, made up from five subjects who were generally familiar with hierarchical control structures but unfamiliar with either monitoring by overhearing or with the evacuation application or its component agents. Each subject was presented with printed books (one for each run) containing the overheard messages (in human-readable form), the same messages overheard by Overseer under optimal (lossless) conditions. As reference material, each subject was given a copy of the plan-hierarchy, team-hierarchy, and the same average duration information available to Overseer (the parameter $\lambda$ for different leaf plans). For each overheard message, a second line of print indicated the time passed since overhearing the message, and the subject was asked to write down their best estimate for the agents' current state (i.e., after the message was overheard and the specified time passed). If they felt different agents or different sub-teams had different states, they were to specify what each agent or subteam is doing. We emphasize that the subjects were presented with exactly the same runs on which Overseer was evaluated.

The actual test process began with a short explanation of the task, with a full explanation of the plan-hierarchy (including answering any questions the subjects had about the semantics of different transitions, etc.), and with a short test run which allowed each subject to use the plan-hierarchy and team hierarchy (but without providing any feedback as to the subject's accuracy). Then, once all questions had been answered, the subjects were presented with the test books and were given unlimited time to complete them, in any order (though once started on a book, they were required to finish it). Subjects were not allowed to look forward at the next message before completing their answer, nor to go back and change their previous answers. They were encouraged to seek assistance in locating plans (in the plan-hierarchy) consistent with messages, but no information was provided as to which plan may be correct if two or more plans were consistent. After all subjects finished with all runs, the test books were taken together and a score was computed. Each correct prediction was awarded a 1, incorrect prediction a 0—same standard used in scoring Overseer's accuracy. A non-specific answer (i.e., "it was somewhere in the sub-tree of `Fly-Flight-Plan`") was considered to be an incorrect prediction, as subjects were repeatedly instructed that a specific answer was required.

The results of the test are presented in Figure 8. As in previous graphs, the Y-axis denotes percentage of accurate monitoring hypotheses across all the data-points in a run (22–45, depending on the run). The X-axis has three categories, for the three different runs. For each run, the left bar (marked "Novices") presents the average accuracy achieved by the novice monitors, the middle bar (marked "Experts") presents the accuracy achieved by the human expert monitor, and the final bar ("YOYO*") re-prints the results presented in Figure 5 above. The results show that the average accuracy of the novices is clearly inferior to that of the expert monitor and to Overseer. Overseer's performance is above that of the human expert in runs I and J. However, the human expert does much better than Overseer in run E.

We draw several conclusions from these results. First, the monitoring task Overseer faced in the evacuation application is not trivial: The novices failed to achieve more than

2. We have had to settle for one expert since training an expert in this task is very time consuming and requires much familiarity with the internals of the evacuation application as well as the TEAMCORE architecture.

70% on average (in their best run), and generally performed significantly worse (by 15% and more) than a human expert. Second, Overseer's performance in different runs was comparable to that of the human expert (sometimes better, sometimes worse). However, Overseer's performance tended to follow the same trend as the novices. In other words, Overseer's accuracy tended to go up and down on different runs in a similar manner to that of the average novice human monitor, while the expert's accuracy remained fairly constant across all runs.

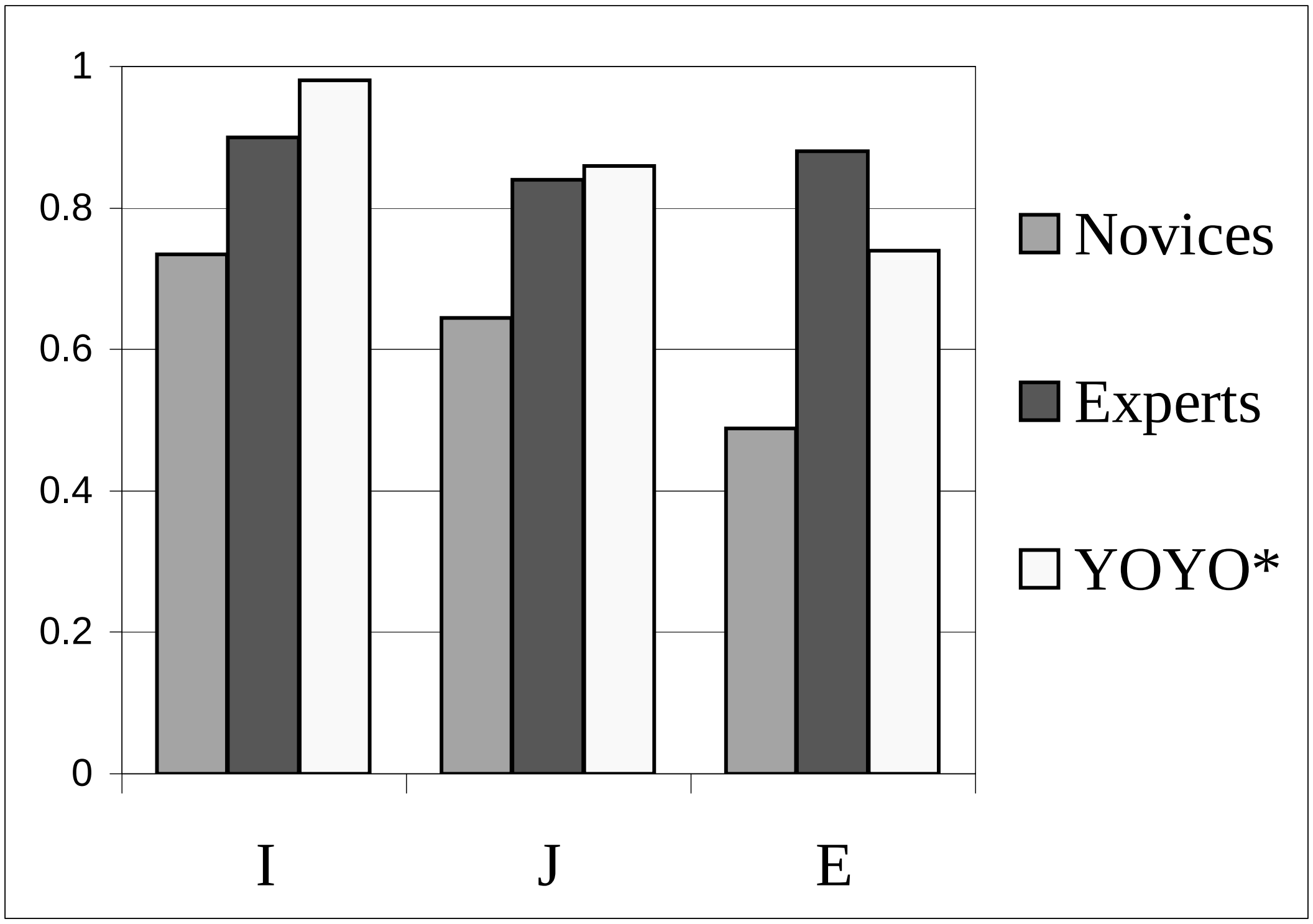

Figure 8: Accuracy of human novice and expert monitors compared to Overseer.

### 6.4 Evaluating YOYO*'s Trading of Expressivity for Scalability

We examine a key trade-off between the expressivity and efficiency involved in the plan-recognition techniques we have presented. From the accuracy discussion above, it is clear that coherence is a useful heuristic. YOYO* takes an extreme approach, strictly ruling out reasoning about incoherences. It is impossible for YOYO*, for instance, to represent an incoherence in which two team-members are in disagreement about the plan executed by the common team. It may thus be impossible for YOYO* to explicitly represent hypotheses associated with communication losses and delays, which cause such incoherences. An approach in which each individual is represented separately allows for such representation, and in this respect is more expressive. However, with a few failure-checks in place, *YOYO* is able to detect many incoherences*, as previously discussed.

On the other hand, YOYO* offers significant computational scalability with respect to the number of agents monitored. Analysis of YOYO*'s complexity (in contrast to the array approach) was already presented in Section 5.2, and we follow it here with empirical evaluation. Figure 9 reports on the space requirement of YOYO* and the array-based approach in three different domains: the evacuation domain, where YOYO* has been evaluated and deployed, and two additional domains in which we have built multi-agent teams—ModSAF (Tambe et al., 1995; Calder, Smith, Courtemanche, Mar, & Ceranowicz, 1993) and RoboCup (Tambe, Adibi, Al-Onaizan, Erdem, Kaminka, Marsella, & Muslea, 1999; Marsella, Adibi, Al-Onaizan, Kaminka, Muslea, Tallis, & Tambe, 2001). YOYO* is currently being evaluated in these domains, and while it has not yet been fully deployed there, we believe the partial existing implementations are sufficient to provide robust projections of the space savings achieved in these domains. We believe that such projected savings of implementation in these two domains could provide a rough guide as to the savings that designers could expect from deploying YOYO* in additional domains.

For each domain, Figure 9 compares the space requirements of the array-based approach (left bar) with those of YOYO* (right bar). In addition, the dark-shaded region on top of each bar shows the space required for representing each additional agent in the two approaches, under the assumption that no additional plans are added to the plan-hierarchy as more agents are added. As discussed above, this assumption is favorable to the array-based representation. The figure shows the significant space savings achieved by YOYO*. First, in representing the teams in their current size, YOYO*'s space requirements are significantly smaller. Furthermore, YOYO*'s savings really shine when we examine the scalability of the two approaches. While the array-based approach requires at least the amount of space shown in the figure as darkly-shaded area, YOYO*'s requirements grow by one node with each additional agent. Its space requirements for representing additional agents are so small, that they don't show in the figure.

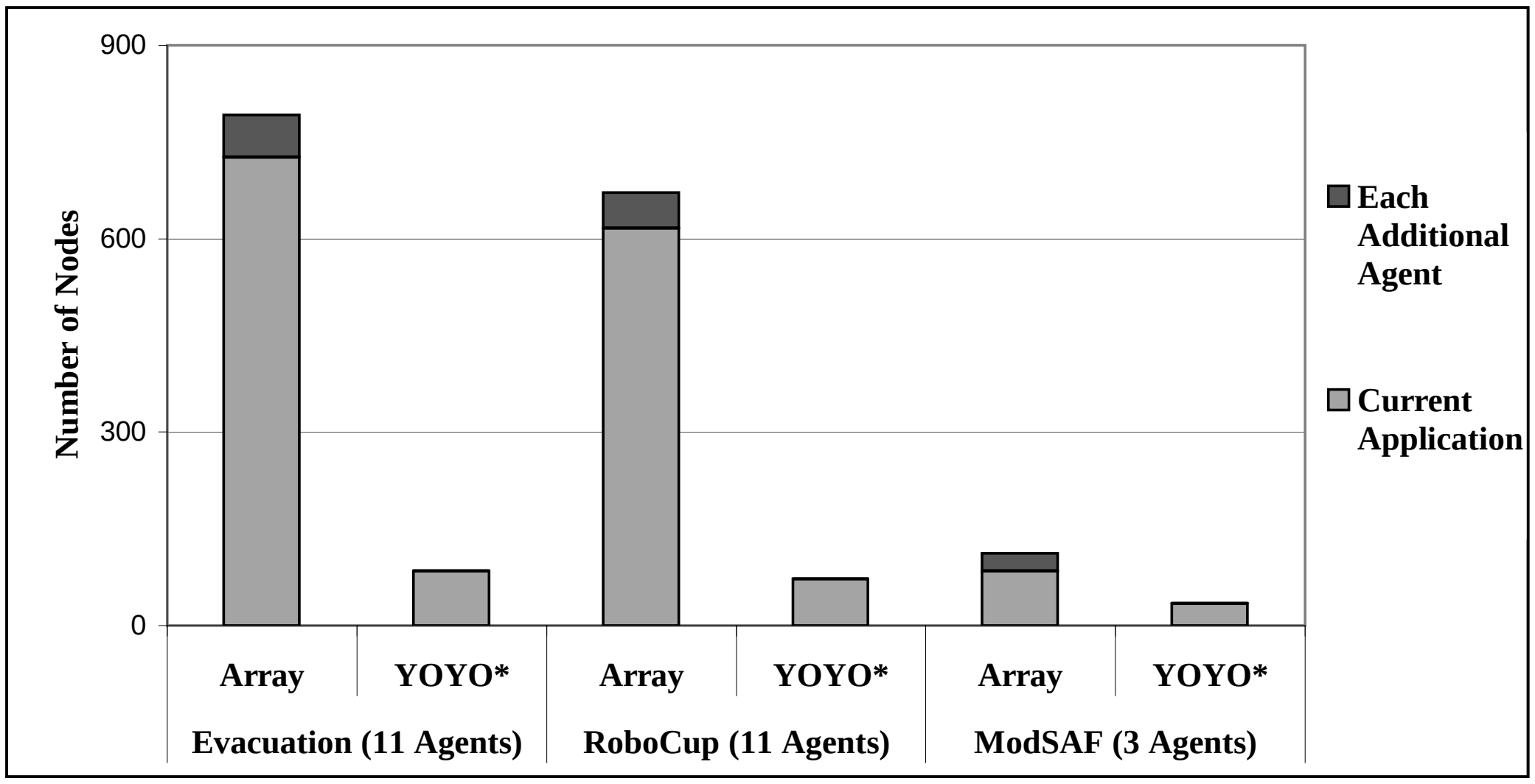

Figure 9: Empirical savings in applying YOYO* in the evacuation and other domains.

Earlier, in Section 5.2, we have analyzed YOYO*'s worst case run-time complexity, but argued that this worst case behavior is very extreme, and cannot be sustained in practice since it involves continuous communications among all agents, the infeasibility of which provided the motivation for exploring a plan-recognition approach. As further evidence for the average case, consider the evacuation application, where agents communicate on average once every 20 state changes. In this application, agents communicate in parallel in 4 or 5 exchanges (out of dozens), but in all cases but one, such parallel communications all referred to the same plan, thus still requiring only a single update in YOYO* (see discussion in Section 5.2). Only once during task execution would 3 agents (out of 11) be expected to communicate in parallel about different plans, a scenario still different than YOYO*'s worst case scenario.

The average length of task execution in this domain is approximately 900 time-ticks. The array approach would update the state of each agent, at each time tick, whether a message would appear or not. Thus its average complexity per time-tick is the same as its worst-case, which is at least $O(MN^2)$. For YOYO*, the average complexity would be significantly different: 899 out of 900 time-ticks it would result in an $O(M+H)$ process, and only one time (out of 900) it would be result in a process three times as expensive (updating the state of 3 different agents). The worst case scenario did not occur at all in any of the different runs.

## 7. Related Work

Aiello et al. (2001) present several benefits to overhearing agent conversations. They suggest that the overhearer may infer the intent of the agents engaged in conversations, and offer specific suggestions for improving the agents' performance. For instance, overhearing a conversation between two agents about a keyword search on the web, the overhearer may suggest alternative keywords to conduct the same search. This work is closely related to our research on Overseer, and indeed points out several potential additional benefits of overhearing technology. However, in contrast to our work, Aiello et al. do not address the problem of intent- or plan-recognition. They do not present algorithms for inferring plans, nor for disambiguating recognized plans.

Overseer differs from most previous work on plan-recognition in being focused on monitoring multiple agents, not a single agent. While previous work in multi-agent plan recognition has either focused on exploiting explicit teamwork reasoning (e.g., Tambe, 1996), or explicitly reasoning about uncertainty when recognizing multi-agent plans (e.g., Devaney & Ram, 1998; Intille & Bobick, 1999), a key novelty in Overseer is that it effectively blends these two threads together. We provide a detailed discussion below.

Like Overseer, $RESC_{team}$ (Tambe, 1996) reasons explicitly about team intentions for inferring *team plans* from observations, similarly to Overseer's use of the coherence heuristic. $RESC_{team}$ uses coherence to restrict the space requirements of the plan-library used, similarly to YOYO*. However, Overseer uses a more advanced teamwork model (e.g., it can predict failure states and recovery actions), uses knowledge about procedures used by a team (i.e., communication decisions), and also explicitly reasons about uncertainty and time, allowing it to answer queries related to the likelihood of current and future team plans (issues not addressed in $RESC_{team}$). Indeed, $RESC_{team}$ does not explicitly represent ordering constraints between plans, and does not address scarce observations: It assumes that observations are available that account for possible changes in the state of each of the observed agents.

Work such as (Devaney & Ram, 1998; Intille & Bobick, 1999) focuses on explicitly addressing uncertainty in plan recognition in multi-agent contexts, but does not exploit explicit notions of teamwork. Devaney and Ram (1998) use pattern matching to recognize team-tactics in military operations. Their approach relies on team-plan libraries, verified by domain experts, that combine the team- and plan-hierarchies; the organizational knowledge is not explicitly represented in their technique. Similarly, Intille and Bobick (1999) rely entirely on coordination constraints among agents to recognize team-tactics in football, and in this sense use a socially-attentive technique that prefers hypotheses in which agents are maintaining their roles. Intille and Bobick's work uses a single structure for each different recognized tactic. Both investigations use position trace data of the monitored human teams.

Our work differs from (Devaney & Ram, 1998; Intille & Bobick, 1999) in several ways. First, these previous investigations have been applied in settings where observations are continuously available about each monitored agent. In contrast, Overseer is targeted towards *overhearing*, where limited observations are available, both in time, and in the number of agents actually observed. Overseer introduces a number of novel techniques (such as the communications predictions) which are useful in such settings. A second important difference is the underlying representation used in reasoning. We introduce a novel representation particularly suited for monitoring by overhearing, while Intille and Bobick rely on standard belief networks, constructed in a particular way to support reasoning about spatial/temporal coordination. Finally, the explicit use we make of teamwork and organizational structure (the team-hierarchy) enables YOYO* in principle to reason about coordination and teamwork failures, where the previous monitoring techniques would fail to recognize the team's actions (Intille & Bobick, 1999).

Huber (1996) reports on the use of probabilistic plan recognition in service of observation-based coordination in the Net-trek domain, and shows that agents using plan recognition for coordination outperform agents using communications for coordination. Huber takes coordination to be cooperative actions on the part of the self- interested agents, e.g., joining an agent in attacking a common enemy. Huber's work does not exploit any knowledge of relationships between the agents to limit the computation or increase the accuracy. Huber's system does allow for some uncertainty caused by missing observations, but in contrast to our work, does not introduce specialized mechanisms (such as ours) to explicitly address these.

Plan Recognition Bayesian Networks (PRBNs) (Charniak & Goldman, 1993) provide a very general model for plan events, evidence, and inference. However, a PRBN is a static Bayesian network, so it must include nodes for all plans and observations throughout the execution of the plans. Therefore, instead of representing only the events of a single time step (as in the DBNs described in Section 3.1), it must include nodes over all time steps. Therefore, for $N$ agents, executing a plan hierarchy of size $M$, over a finite time horizon of $T$ steps, the number of nodes in the network will be $O(TNM^2)$. Inference will have a space/time complexity exponential in the number of nodes, $O(2^{TNM^2})$, which is prohibitive over the lengths of execution found in our example domains (e.g., $T = 900$).

The representation used by YOYO* is related to existing approaches to the modeling of stochastic processes, in particular those used for probabilistic plan recognition. The representation we present perhaps most closely resembles Hidden Markov Models (HMMs) (Rabiner, 1989), used for plan-recognition in (Han & Veloso, 1999). One could, in theory, represent the plan state of a team of agents within the unconstrained state space of an HMM. However, the HMM state space would have to represent all possible combinations of the individual plan states of the agents, so the size of the HMM state space would be exponential in the number of agents and plans. Thus, the standard algorithms for HMM inference would not be able to exploit the structure of the plan and team hierarchies, nor the particular forms of evidence (as described in Section 3.2), in the way that we do in YOYO*. Generalized versions of the HMM model (Ghahramani & Jordan, 1997; Jordan, Ghahramani, & Saul, 1997) could more compactly represent the same state space as in YOYO*, but exact inference is intractable for these models. These models have more efficient algorithms for approximate inference, but these would have difficulty with the determinism present in our planning models.

Pynadath and Wellman report on the Probabilistic State-Dependent Grammar (PSDG) model (2000) that avoids the full complexity of DBN inference by making simplifying assumptions appropriate for plan recognition. However, while PSDG can incorporate broader classes of inference than YOYO*, it is intended for single-agent plan recognition, and does not support concurrency in a general enough fashion for multi-agent plan recognition.

Goldman, Geib and Miller (1999) develop a conceptual model for Bayesian plan recognition which does include, as one of its key novelties, the ability to infer the plans of a single agent from lack of observation of its action. However, Goldman et al. deal with a different issue altogether than the one our communications predictions address. Their framework looks at a sequence of observations, in which an observation may be missing, but observations of actions following it appear. Their framework then allows inference that plans that should have given rise to the missing observation can be ruled out as recognition hypotheses. In contrast, our approach uses the communications predictions to make inference of plan-steps that did not *yet* occur. Overseer probabilistically expects the predictions to come true, and does not infer additional information from a missing (predicted) observation that is followed by another. In addition, our approach is fully implemented and deployed in multi-agent settings, rather than single agent.

A complementary line of work (in the context of the TEAMCORE architecture) has focused on *intended* plan-recognition for monitoring, where team-members may adapt their communications such that monitoring is made easier (Tambe et al., 2000). This work (i) reduced, but did not eliminate uncertainty, and (ii) did not present any methods to address uncertainty, as we do here, However, it presents an interesting future direction for Overseer's development.

## 8. Summary and Future Work

This paper introduced monitoring by overhearing, a technique that will be increasingly important with the growing need to monitor agent systems, particularly distributed or deployed. We presented Overseer, a system for monitoring teams by overhearing the routine communications team-members exchange as part of the execution of their joint tasks. Monitoring by overhearing, while being a plan-recognition task, presents characteristic challenges not previously addressed. These include the scarcity of observations compared to the rate of change in agent's state, and the fact that agents are not individually observable, as the observations are essentially of multi-agent actions. In addition to these, familiar challenges such as demanding response times and maintaining performance in face of a scale-up in the number of monitored agents, are also present.

To address these challenges, Overseer employs a number of novel techniques, which exploit knowledge of the relationships between the agents to alleviate uncertainty and increase efficiency of monitoring: (i) An efficient probabilistic algorithm for plan-recognition, particularly suited for monitoring communications; (ii) YOYO*, an approach for efficient maintenance of recognition of coherent hypotheses; and (iii) use of social structures and procedures, e.g., team coherence and communications to maintain coherence, to alleviate uncertainty. To demonstrate the generality of these techniques, we have discussed the potential use of these techniques with representations other than a plan-hierarchy, in particular DBNs (Kjærulff, 1992).

We provided an in-depth empirical evaluation of these techniques in one of the domains in which Overseer is applied. The evaluation carefully examines the contribution of each technique to the overall recognition success, and demonstrates that these techniques work best together, as they complement relative weaknesses of each other. The paper also presented an evaluation of the scalability of YOYO*, and its performance under conditions of observation loss. Finally, we presented a comparison of Overseer's performance with that of human expert and novice monitors, and demonstrated that Overseer performance is comparable to that of human experts, despite the difficulty of the monitoring task.

Several opportunities for future research directions arise from the experimental results. First, the use of rote-learning to predict when messages will be observed (provided as feasibility demonstration), proved effective for normative runs. However, the simple mechanism was damaging when rare patterns of communications arose, as some of the experiments have shown. In-depth exploration of the role of learning is therefore one of the directions we hope to pursue in the future. In addition, learning mechanisms that can derive plan-hierarchy and team-hierarchy structures from records of conversations are also of much interest.

## Acknowledgements

This paper is based in part on an Agents-2001 paper by the same authors (Kaminka, Pynadath, & Tambe, 2001). Parts of this research were carried out while the first author was a Post Doctorate Fellow at the Computer Science Department, Carnegie Mellon University. We thank Manuela Veloso for her enthusiastic support of this project at Carnegie Mellon University, and we thank Yves Lespérance, Victor Lesser, George Bekey, Jeff Rickel, and Dan O'Leary for useful comments. Oshra Kaminka deserves special thanks for her help in analyzing and processing the data. This research was supported by DARPA awards F30602-98-2-0108, F30602-98-2-0135, and F30602-00-2-0549, managed by the Air Force Research Labs/Rome site.

## Appendix A. Additional algorithms and proofs

This appendix contains the pseudo-code for all algorithms described in the paper, for which pseudo-code was not provided in the body of the text itself. These include the modifications to the propagation procedures necessary for propagation within YOYO*. In addition, we provide a proof that the number of coherent hypotheses for $N$ agents is linear in the size of the plan-library $M$.

### A.1 The Number of Incoherent and Coherent Hypotheses

Let $M_i$ be the monitoring plan-library for agent $i, 1 \leq i \leq N$. When monitoring agent $i$, a monitoring system reasons about monitoring hypotheses in $M_i$. In other words, we can view $M_i$ as the finite set of all possible plans agent $i$ may be executing. Given a query as to the agent's current state by the monitoring system, the plan-recognition algorithm picks some $k_i$ specific members of $M_i$ as hypotheses as to the current state of the agent—call these sets of hypotheses $m_i$ where $|m_i| = k_i$.

To construct an overall team hypothesis, the monitoring system must combine the individual hypotheses to form a hypothesis for the team's state. For each agent $i$, the monitoring system chooses one individual hypothesis $h_i \in m_i$. The combination of these forms the team state hypothesis. If there is no uncertainty about the state of any of the agent, i.e., $k_i = 1$ for all $i$, then one team hypothesis exists. However, if uncertainty exists about the state agents, then clearly, the process of selecting individual hypotheses becomes combinatorial in nature, as all possible combinations of all individual hypotheses are possible in principle.

Let us consider how many coherent hypotheses exist. If we restrict ourselves to coherent hypotheses, then the selection of individual hypotheses for each agent are constrained such that the selections are in agreement—the same individual hypothesis is selected for each agent. Given a selection of an individual state hypothesis $h_1 \in m_1$ for the first agent, we must choose $h_2 \in m_2$ for the second agent, $h_3 \in m_3$ for the third agent, etc., such that $h_1 = h_2 = h_3 = ... = h_N$. Since there are not more than $k_1 \leq |M_1|$ individual state hypotheses for the first agent, it follows that the number of coherent team-state hypotheses is bounded by $|M_1|$, i.e., the size of the plan library for the agents. In fact, the number of coherent hypotheses is bounded by $\min k_i$ —since only members of $m_{\min k_i}$ can be matched with members of the other individual hypothesis sets, $m$. In contrast, by definition, all other team-state hypotheses are incoherent. There will be $k_1 * k_2 * k_3 * ... * k_N - (\min k_i)$ of these hypotheses.

### A.2 YOYO* Propagation Algorithms (Section 5.1)

The algorithms presented in this section support those presented in the main text of the paper, and are provided here for completeness. Some of them may contain a step which iterates over all teams that can take an outgoing transition (e.g., line 1 of algorithm 6, or line 13 of algorithm 7). This step requires some further clarification: When iterating over all outgoing teams that meet the condition, the algorithm consults the team-hierarchy to carry out the iteration only for the *topmost* teams (in terms of the team-hierarchy) that meet the condition. For instance, in our application domain, the team TASK-FORCE has (among others) two subteams TRANSPORTS and ESCORTS. If a transition is allowed to be taken by TRANSPORTS only, then an iteration "over all teams that are allowed to take the transition" will not consider either ESCORTS or TASK-FORCE. However, if the transition allows TASK-FORCE, then the iteration step will take place only once—it will be executed once for the team TASK-FORCE, which is the parent team for TRANSPORTS and ESCORTS.

**Algorithm 6** TEAM-PROPAGATE-DOWN(**plan** $Y$, **probability** $\rho$, **beliefs**, $b$, **plans** $M$)

**for all** teams $T$ who are allowed to take an outgoing hierarchical-decomposition transition from $Y$ **do**
    $C_T \leftarrow \{c \mid c \in M,\ c \text{ first child of } Y,\ \text{c is to be taken by team } T\}$
    $\rho' \leftarrow \rho / \mid C_T \mid$
    **for all** plans $c \in C_T$ **do**
        $b_{t+1}(Y, \neg block) \leftarrow b_{t+1}(Y, \neg block) + \rho'$
        TEAM-PROPAGATE-DOWN($c, \rho', b, M$)

**Algorithm 7** TEAM-PROPAGATE-FORWARD(**team-hierarchy** $H$, **beliefs** $b$, **plans** $M$)

1: **for all** plans $X \in M$ **do**
2:     $b_{t+1}(X, \neg block) \leftarrow 0.0$
3:     $b_{t+1}(X, block) \leftarrow 0.0$
4:     $out_x \leftarrow 0.0$
5:     $\eta_x \leftarrow 0.0$
6: **for all** plans $X \in M$ in *post-order* **do** {children in temporal order before parents}
7:     **if** X is a leaf **then**
8:         $out_x \leftarrow b_t(X, \neg block)(1 - e^{-\lambda_x})$ {calculate probability of $X$ terminating at time $t$}
9:     **else** {X is a parent}
10:         $out_x$ is known { because post-order guarantees all children set it in line 21}
11:     **for all** *temporal* outgoing transitions $T_{x \to y}$ from $X$ **do**
12:         $\eta_x \leftarrow \eta_x + (1 - \mu_{xy})\pi_{xy}$
13:     **for all** teams $E$ who are allowed to take a temporal outgoing transition **do**
14:         **if** $\eta_x > 0$ **then** {some transition can be taken}
15:             **for all** temporal outgoing transitions $T_{x \to y}$ from $X$ to be taken by $E$ **do**
16:                 $\rho \leftarrow out_x(1 - \mu_{xy})\pi_{xy}$
17:                 **if** $T_{x \to y}$ leads to a successor plan $Y$ **then**
18:                     $b_{t+1}(Y, \neg block) \leftarrow b_{t+1}(Y, \neg block) + \rho$
19:                     TEAM-PROPAGATE-DOWN($Y, \rho, b, M$)
20:                 **else** {$T_{x \to y}$ is a terminating transition}
21:                     $out_{parent(x)} \leftarrow out_{parent(x)} + (1 - \mu_{xy})\pi_{xy}$ {parent's outgoing probability is its children's}
22:     $b_{t+1}(X, block) \leftarrow b_{t+1}(X, block) + out_x - \eta_x$
23:     $b_{t+1}(X, \neg block) \leftarrow b_{t+1}(X, \neg block) - out_x$

Algorithm 8 below may require some clarifications. First, it is important to note that the plans $Y$ (line 1) are traversed in pre-order—parents before children. The scaling calculation depends on the parent having the scaled probability. Second, the iteration over sub-plans $Y$ essentially captures all plans in the subtree rooted in the parent plan $P$, except for those in the subtree rooted by $P$'s child $X$, which already has been adjusted by YOYO* prior to the call to this algorithm. In fact, the use of $X$'s team $T$ to scale only other plans makes sure that any of $X$'s siblings, that are alternatives to $X$ for the team $T$, do not get scaled. This is correct because this procedure is called when incorporating evidence for $X$ (rather than any of its siblings).

**Algorithm 8** Scale(**parent plan** $P$, **team** $T$, **child plan** $X$, **beliefs** $b$)

1: **for all** subplans $Y$ of $P$, where $team(Y) \neq T$, in *pre-order* **do**

2: $b_{t+1}(Y, \neg block) \leftarrow b_{t+1}(Y, \neg block) + \frac{b_t(Y, \neg block)}{b_t(parent(Y), \neg block)} b_{t+1}(parent(Y), \neg block)$

3: $b_{t+1}(Y, block) \leftarrow b_{t+1}(Y, block) + \frac{b_t(Y, block)}{b_t(parent(Y), \neg block)} b_{t+1}(parent(Y), \neg block)$